\documentclass[final]{siamltex}

\usepackage{algorithm, algorithmic}
\usepackage{times}
\usepackage{epsfig}
\usepackage{graphicx}
\usepackage{amsmath,amssymb}
\usepackage{fancybox}
\usepackage{alltt}
\usepackage{soul}
\usepackage{color}
\usepackage{verbatim}
\usepackage{xcolor}
\usepackage{colortbl,hhline}
\usepackage{framed}
\def\bb{{\mathbf b}}
\def\xb{{\mathbf x}}
\def\wb{{\mathbf w}}
\def\vb{{\mathbf v}}

\def\Wb{{\mathbf W}}
\def\yb{{\mathbf y}}
\def\Yb{{\mathbf Y}}
\def\Xb{{\mathbf X}}
\def\Sigmab{{\mathbf \Sigma}}
\def\G{{\mathcal{G}}}
\def\Real{{\mathbb{R}}}
\def\defin{\triangleq}

\graphicspath{{img/}}

\title{Multi-scale Mining of fMRI Data with Hierarchical Structured Sparsity}

\author{Rodolphe Jenatton
\thanks{INRIA Rocquencourt - Sierra Project-Team, Laboratoire d'Informatique de l'Ecole Normale Sup\'erieure, INRIA/ENS/CNRS UMR 8548
(\texttt{firstname.lastname@inria.fr}).}
\and Alexandre Gramfort
\thanks{INRIA Saclay - Parietal Project-Team, CEA Neurospin.
(\texttt{firstname.lastname@inria.fr}).}
\and Vincent Michel$\ ^\dag$
\and Guillaume Obozinski$\ ^*$
\and Evelyn Eger
\thanks{INSERM U562, France - CEA/ DSV/ I2BM/ Neurospin/ Unicog
(\texttt{evelyn.eger@cea.fr}).}
\and Francis Bach$\ ^*$
\and Bertrand Thirion$\ ^\dag$
}

\begin{document}

\maketitle

\begin{abstract}
Reverse inference, or \emph{``brain reading''}, is a recent paradigm for
analyzing functional magnetic resonance imaging (fMRI) data, based on pattern
recognition and statistical learning. By predicting some cognitive variables related to brain
activation maps, this approach aims at decoding brain activity.
Reverse inference takes into account the multivariate information between
voxels and is currently the only way to assess how precisely
some cognitive information is encoded by the activity of neural populations within the whole brain.
However, it
relies on a prediction function that is plagued by the curse of dimensionality,
since there are far more features than samples, \emph{i.e.}, more voxels than fMRI
volumes.
To address this problem, different methods have been proposed, such as, among others, 
 univariate feature selection, feature agglomeration and
regularization techniques.
In this paper, we consider a sparse hierarchical structured regularization.
Specifically, the penalization we use is constructed from a tree that is obtained
by spatially-constrained agglomerative clustering.
This approach encodes the
spatial structure of the data at different scales into the regularization,
which makes the
overall prediction procedure more robust to inter-subject variability.
The regularization used induces the selection of spatially coherent predictive brain regions 
simultaneously at different scales.
We test our algorithm on real data acquired to study the
mental representation of objects, and we show that the proposed algorithm not only delineates meaningful
brain regions but yields as well better prediction accuracy than reference methods.
\end{abstract}



\pagestyle{myheadings}
\thispagestyle{plain}
\markboth{R. Jenatton, A. Gramfort, V. Michel, G. Obozinski, E. Eger, F. Bach and B. Thirion}
{Multi-scale Mining of fMRI Data with Hierarchical Structured Sparsity}

\footnotetext{A preliminary version of this work appeared in~\cite{Jenatton2011}.}
\section{Introduction}
Functional magnetic resonance imaging (or fMRI) is a widely used functional
neuroimaging modality. Modeling and statistical analysis of fMRI data are
commonly done through a linear model, called general linear model (GLM) in the community,
that incorporates
information about the different experimental conditions and the dynamics of the hemodynamic
response in the design matrix. 
The experimental paradigm consists of a sequence of stimuli, \emph{e.g.},
visual and auditory stimuli, which are included as regressors in the design matrix after convolution with a suitable hemodynamic filter.
The resulting model
parameters---one coefficient per voxel and regressor---are known as
\emph{activation maps}. They represent the local influence of the different
experimental conditions on fMRI signals at the level of individual voxels. The
most commonly used approach to analyze these activation maps is
called classical inference. It relies on mass-univariate statistical tests (one
for each voxel), and yields so-called statistical parametric maps
(SPMs)~\cite{friston1995c}. Such maps are useful for functional brain mapping,
but classical inference has some limitations: it suffers from multiple
comparisons issues and it is oblivious of the multivariate structure of fMRI data. Such
data exhibit natural correlations between neighboring voxels forming clusters with different
sizes and shapes, and also between distant but functionally connected brain
regions.

To address these limitations, an approach called reverse inference (or
``brain-reading'')~\cite{dehaene1998,cox2003} was recently proposed. Reverse 
inference relies on pattern recognition tools and statistical learning methods
to explore fMRI data. Based on a set of activation maps, reverse inference
estimates a function that can then be used to predict a target (typically,
a variable representing a perceptual, cognitive or behavioral parameter) for a
new set of images. The challenge is to capture the correlation structure
present in the data in order to improve the accuracy of the fit,
which is measured through the resulting prediction accuracy. Many standard
statistical learning approaches have been used to construct prediction functions, among
them kernel machines (SVM, RVM)~\cite{Scholkopf2002} or discriminant analysis
(LDA, QDA)~\cite{Hastie2009}. For the application considered in this paper, earlier performance
results~\cite{cox2003,laconte2005} indicate that we can restrict ourselves to
mappings that are linear functions of the data.

Throughout the paper, we shall consider a training set composed of $n$ pairs
$(\xb,y) \in \Real^p \times \mathcal{Y}$, where $\xb$ denotes a $p$-dimensional
fMRI signal ($p$ voxels) and $y$ stands for the target we try to predict.
Each fMRI data point $\xb$ will correspond to an activation map after GLM fitting.
In the experiments we carry out in Section~\ref{sec:exp}, we will encounter both
the regression and the multi-class classification settings, where $\mathcal{Y}$ denotes 
respectively the set of real numbers and a finite set of integers.
An example of a regression setting is the prediction of a pain level
from fMRI data~\cite{marquand2010}
or in the context of classification, the prediction of object
categories~\cite{cox2003}. Typical datasets consists of a few hundreds
of measurements defined each on a $2\times2\times2$-mm voxels grid
forming $p \approx 10^5$ voxels when working with full brain data.
Such numbers, given as illustration, are not intrinsic limitation of MRI
technology and are still regularly improved by experts in the field.

In this paper, we aim at learning a weight vector $\wb \in
\Real^p$ and an intercept $b \in \Real$ such that the prediction of $y$ can be
based on the value of $\wb^\top \xb + b$. This is the case for
the linear regression and
logistic regression models that we use in Section~\ref{sec:exp}.
The scalar $b$ is not particularly informative, however the vector $\wb$
corresponds to a volume that can be represented in brain space as a volume
for visualization of the predictive pattern of voxels.
It is useful to rewrite these quantities in
matrix form; more precisely, we denote by $\Xb \in \Real^{n\times p}$ the
design matrix assembled from $n$ fMRI volumes and by $\yb \in \Real^n$ the
corresponding $n$ targets. In other words, each row of $\Xb$ is a
$p$-dimensional sample, \emph{i.e.}, an activation map of $p$ voxels related to
one stimulus presentation.

Learning the parameters $(\wb,b)$ remains challenging since the number of features
($10^4$ to $10^5$ voxels) exceeds by far the number of samples (a few hundreds
of volumes). The prediction function is therefore prone to overfitting in which
the learning set is predicted precisely whereas the algorithm provides very inaccurate predictions on new samples (the
test set). To address this issue, \emph{dimensionality reduction} attempts to find a low dimensional subspace
that concentrates as much of the predictive power of the original set as possible for the problem at hand. 

Feature selection is a natural approach to perform dimensionality reduction in fMRI, since reducing the number of voxels makes it easier to identify a predictive region of the brain.
This corresponds to discarding some columns of $\Xb$.
This feature selection can be univariate, \emph{e.g.}, analysis of variance
(ANOVA)~\cite{Lehmann2005}, or multivariate. While univariate methods ignore joint
information between features, multivariate approaches are more adapted to
reverse inference since they extract predictive patterns from the data as a whole.
However, due to the huge number of possible patterns, these approaches suffer
from combinatorial explosion, and some costly suboptimal heuristics (\emph{e.g.},
recursive feature elimination \cite{guyon2002,DeMartino2008}) can be used.
That is why ANOVA is usually preferred in fMRI. Alternatively, two more adapted
solutions have been proposed: \emph{regularization} and \emph{feature
agglomeration}.

Regularization is a way to encode a priori knowledge about the weight vector
$\wb$. Possible regularizers can promote for example spatial smoothness or sparsity
which is a natural assumption for fMRI data. Indeed, only a few brain regions are
assumed to be significantly activated during a cognitive task. Previous
contributions on fMRI-based reverse inference include
\cite{carroll2009,rissman2010,ryali2010,yamashita2008}. They can be presented
through the following minimization problem:
\begin{equation}
\min_{(\wb,b) \in \Real^{p+1}} \mathcal{L}(\yb,\Xb,\wb,b) + \lambda \Omega(\wb)\quad
\text{with}\quad \lambda \geq 0,
\label{eq:opt_pb}
\end{equation}
where $\lambda \Omega(\wb)$ is the regularization term, typically a
non-Euclidean norm, and the fit to the data is measured through a convex loss
function $(\wb,b)\mapsto\mathcal{L}(\yb,\Xb,\wb,b) \in \Real_+$. The choice of
the loss function will be made more specific and formal in the next sections.
The coefficient of regularization~$\lambda$ balances the loss and the penalization term. In this
notation, a common regularization term in reverse inference is the so-called \emph{Elastic
net}~\cite{zou2005,Grosenick2009}, which is a combined $\ell_{1}$ and
$\ell_{2}$ penalization:
\begin{equation}
\lambda \Omega(\wb) = \lambda_1 \|\wb\|_{1}  + \lambda_2
\|\wb\|^{2}_{2} = \sum_{j = 1}^{p}\big\{ \lambda_1 |\wb_j| + \lambda_2 \wb_j^2\big\}.
\end{equation}
For the squared loss, when setting $\lambda_1$ to 0, the model is called ridge regression, while when
$\lambda_2 = 0$ it is known as Lasso~\cite{Tibshirani1996} or basis
pursuit denoising (BPDN)~\cite{Chen1998}. The essential shortcoming of the Elastic net is that it
does not take into account the spatial structure of the data, which is crucial
in this context~\cite{michel-etal:11}. Indeed, due to the intrinsic smoothing of the complex
metabolic pathway underlying the difference of blood oxygenation measured with fMRI~\cite{ugurbil2003},
statistical learning approaches should be informed by the 3D grid structure
of the data.

In order to achieve dimensionality reduction, while taking into account the spatial
structure of the data, one can resort to \emph{feature agglomeration}. 
Precisely, new features called \emph{parcels} are naturally generated via averaging of groups of neighboring voxels
exhibiting similar activations.
The advantage of agglomeration is that no information is discarded a priori
and that it is reasonable to hope that averaging might reduce noise. 
Although, this approach has been successfully
used in previous work for brain mapping~\cite{flandin2002,thirion2006}, existing work
does typically not consider the supervised information (\emph{i.e.}, the target
$y$) while exploring the parcels. A recent approach has been proposed to
address this issue, based on a supervised greedy top-down exploration of a tree obtained
by hierarchical clustering~\cite{michel2010}. 
This greedy approach has proven
to be effective especially for inter-subject analyzes, \emph{i.e.}, when
training and evaluation sets are related to different subjects.
In this context,
methods need to be robust to intrinsic spatial variations that exist
across subjects:
although a preliminary co-registration to a common space has been performed,
some variability remains between subjects,
which implies that there is no
perfect voxel-to-voxel correspondence between volumes. As a result, the
performances of traditional voxel-based methods are strongly affected. 
Therefore, averaging in the form of parcels is a good way to cope with
inter-subject variability. This greedy approach is nonetheless suboptimal, 
as it explores only a subpart of the whole tree.

Based on these considerations, we propose to integrate the multi-scale spatial
structure of the data \textit{within} the regularization term $\Omega$, while
preserving convexity in the optimization. This notably guarantees global
optimality and stability of the obtained solutions. To this end, we design a
sparsity-inducing penalty that is directly built from the hierarchical
structure of the spatial model obtained by Ward's algorithm~\cite{ward1963}
using a contiguity-constraint~\cite{murtagh85}.
This kind of penalty has already been successfully applied in several contexts,
\emph{e.g.}, in bioinformatics, to exploit the tree structure of gene networks
for multi-task regression~\cite{Kim2009}, in log-linear models for the selection
of potential orders~\cite{Schmidt2010}, in image processing for wavelet-based denoising~\cite{Baraniuk2008,Jenatton2010b,Rao2011}, 
and also for topic models~\cite{Jenatton2010b}.
Other applications have emerged in natural language~\cite{Martins2011} and audio processing~\cite{Sprechmann2010a}. 

We summarize here the contributions of our paper:
\begin{itemize}
  \item We explain how the multi-scale spatial structure of fMRI data can be taken into account in the context of reverse inference through the combination of a spatially constrained hierarchical clustering procedure and a sparse hierarchical regularization.
  \item We provide a convex formulation of the problem and propose an efficient optimization procedure.
  \item We conduct an experimental comparison of several algorithms and formulations on fMRI data and illustrate the ability of the proposed method to localize in space and in scale some brain regions involved in the processing of visual stimuli.
\end{itemize}

The rest of the paper is organized as follows: we first present the concept of
structured sparsity-inducing regularization and then describe the different
regression/classification formulations we are interested in. After exposing how
we handle the resulting large-scale convex optimization problems thanks to
a particular instance of proximal methods---the forward-backward splitting algorithm, we validate our approach both in a synthetic setting and on a
real dataset.

\section{Combining agglomerative clustering with sparsity-inducing regularizers}

As suggested in the introduction, it is possible to construct a tree-structured hierarchy of new features on top of the original voxels using hierarchical clustering.  
Moreover, spatial constraints can be enforced in the clustering algorithm so that the underlying voxels corresponding to each of these 
features form localized spatial patterns on the brain similar to the ones we hope to retrieve~\cite{chklovskii2004}.
Once these features constructed, instead of selecting features in the tree greedily,
we propose to cast the feature selection problem as supervised learning problem of the form
~(\ref{eq:opt_pb}). One of the qualities of the greedy approach however is that it is only allowed to select potentially more noisy features, corresponding to 
smaller clusters, after the a priori more stable features associated with ancestral clusters in the hierarchy have been selected.
As we will show, it is possible to construct a convex regularizer $\Omega$ that has the same property, i.e. that respects the hierarchy, and prioritizes
the selection of features in the same way. Naturally, the regularizer has to be constructed directly from the hierarchical clustering of the voxels. 

\subsection{Spatially-constrained hierarchical clustering}

Starting from $n$ fMRI volumes $\Xb=[\xb^1,\dots,\xb^p] \in \Real^{n\times p}$ described by $p$ voxels, 
we seek to cluster these voxels so as to produce a hierarchical representation of $\Xb$. 

\begin{figure}[h!]
\centering
\hfill
\begin{minipage}{0.45\columnwidth}
    \caption{
    Example of a tree $\mathcal{T}$ when $p=5$, with three voxels and
    two parcels. The parcel 2 is defined as the averaged intensity of the
    voxels $\{1,2\}$, while the parcel 1 is obtained by averaging the values of the voxels $\{1,2,3\}$.
    In red dashed lines are represented the five groups of
    variables that compose $\G$. For instance, if the group containing the
    parcel 2 is set to zero, the voxels $\{1,2\}$ are also (and necessarily)
    zeroed out. Best seen in color.}
    \label{fig:tree}
\end{minipage}
\hfill
\begin{minipage}{0.45\columnwidth}
    \includegraphics[width=0.9\linewidth]{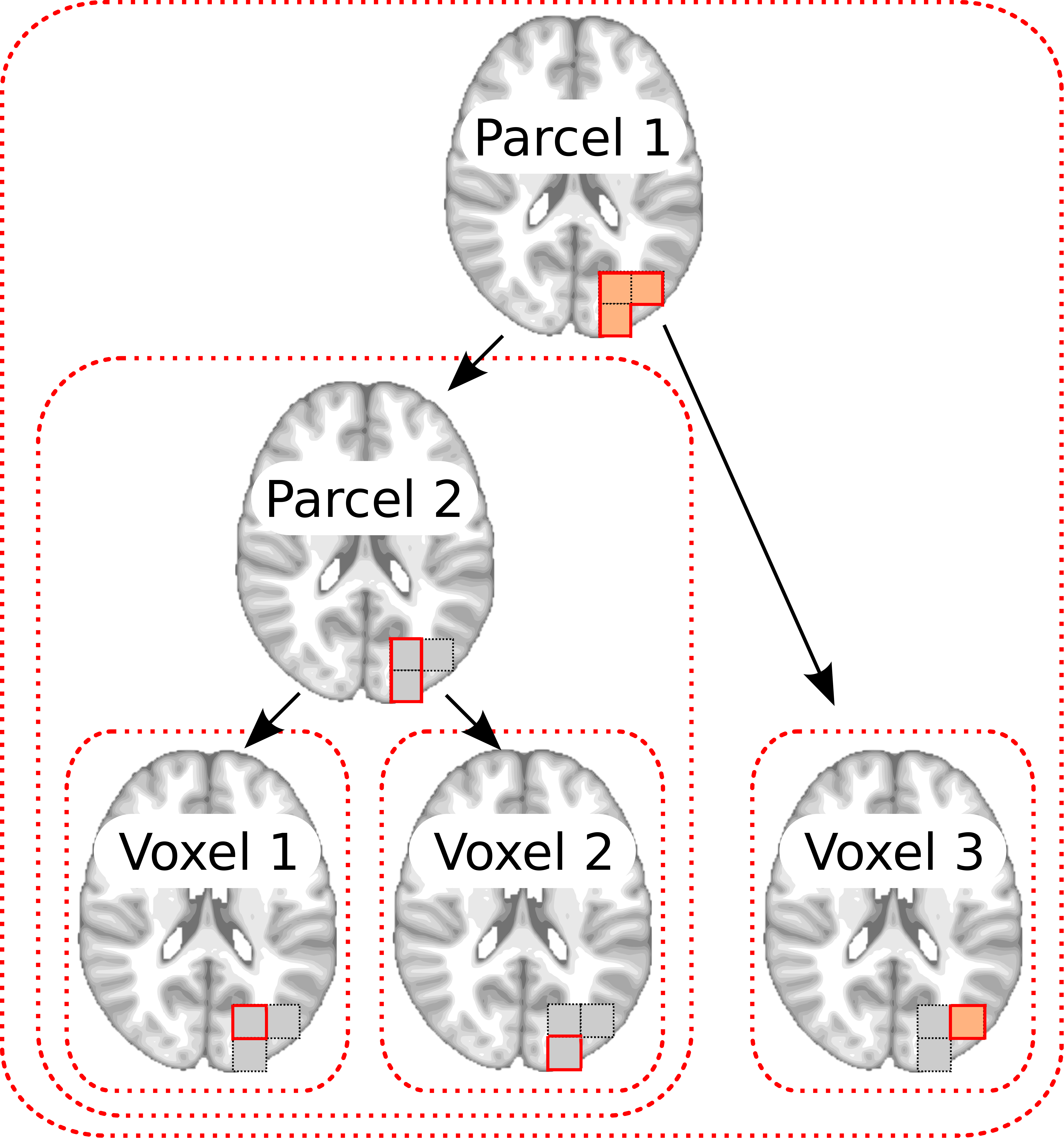}
\end{minipage}
\hfill
\end{figure}

To this end, we consider \emph{hierarchical agglomerative clustering} procedures~\cite{johnson1967}.
These begin with every voxel $\xb^j$ representing a singleton cluster $\{j\}$, 
and at each iteration, a selected pair of clusters---according to a criterion discussed below---is merged into a single cluster. 
This procedure yields a hierarchy of
clusters represented as a binary tree $\mathcal{T}$ (also often called a
dendrogram)~\cite{johnson1967}, where each nonterminal node is associated with the cluster obtained by merging its two children clusters.
Moreover, the root of the tree $\mathcal{T}$ is the unique cluster that
gathers all the voxels, while the leaves are the clusters consisting of a single voxel.
From now on, we refer to each nonterminal node of $\mathcal{T}$ as a \emph{parcel},
which is the union of its children's voxels (see Figure~\ref{fig:tree}).

Among different hierarchical agglomerative clustering procedures, we use
the variance-minimizing approach of Ward's algorithm~\cite{ward1963}. 
In short, two clusters are merged if the resulting cluster minimizes the sum of
squared differences of the fMRI signal within all clusters (also known as \emph{inertia criterion}).
More formally, at each step of the procedure, 
we merge the clusters $c_1$ and $c_2$ that minimize the criterion
\begin{eqnarray}
\Delta(c_1,c_2) &=& \sum_{ j \in c_1 \cup c_2   }\| \xb^j  - \langle  \Xb \rangle_{ c_1 \cup c_2  }  \|_2^2  - 
\Big(   \sum_{ j \in c_1   }\| \xb^j  - \langle  \Xb \rangle_{ c_1 }  \|_2^2 + \sum_{ k \in c_2   }\| \xb^k  - \langle  \Xb \rangle_{ c_2 }  \|_2^2     \Big)\notag \\
&=&  \frac{ |c_1||c_2|  }{ |c_1|+|c_2| } \| \langle  \Xb \rangle_{ c_1 } - \langle  \Xb \rangle_{ c_2 }   \|_2^2 \label{eq:ward_criterion}, 
\end{eqnarray}
where we have introduced the average vector $\langle  \Xb \rangle_{ c } \defin \frac{1}{|c|} \sum_{ j\in c } \xb^j $.
In order to take into account the spatial information, we also add connectivity
constraints in the hierarchical clustering algorithm, so that only neighboring
clusters can be merged together. 
In other words, we try to minimize the criterion $\Delta(c_1, c_2)$ 
only for pairs of clusters which share neighboring voxels (see Algorithm~\ref{alg:ward}). 
This connectedness constraint is important since the resulting clustering is likely to differ from 
 standard Ward's hierarchical clustering.
\subsubsection{Augmented space of features}
Based on the output of the hierarchical clustering presented previously, 
we define the following augmented space of variables (or \textit{features}):
instead of representing the $n$ fMRI volumes only by its individual voxel intensities, we add to it a vector with levels of activation of each of the parcels at the interior nodes
of the tree $\mathcal{T}$, which we obtained from the agglomerative clustering algorithm.
Since $\mathcal{T}$ has $q\defin|\mathcal{T}|=2p-1$ nodes,\footnote{We can then identify nodes (and parcels) of
$\mathcal{T}$ with indices in $\{1,\dots,q\}$.}, the data obtained in the augmented space can be gathered in a matrix $\tilde{\Xb} \in \Real^{n\times q}$.

In the following, the level of
activation of each parcel is simply the averaged intensity of
the voxels it is composed of (\emph{i.e.}, local
averages)~\cite{flandin2002,thirion2006}. This produces a multi-scale
representation of the fMRI data that has the advantage of becoming increasingly invariant to spatial
shifts of the encoding regions within the brain volume.
We summarize the procedure to build the enlarged matrix $\tilde{\Xb} \in \Real^{n\times q}$ in Algorithm~\ref{alg:ward}.
\begin{algorithm}
   \caption{Spatially-constrained agglomerative clustering and augmented feature space.}
   \label{alg:ward}
\begin{algorithmic}
   \STATE {\bfseries Input:} $n$ fMRI volumes $\Xb=[\xb^1,\dots,\xb^p] \in \Real^{n\times p}$ described by $p$ voxels.
   \STATE {\bfseries Output:} $n$ fMRI volumes $\tilde{\Xb} \in \Real^{n\times q}$ in the augmented feature space.
   \STATE {\bfseries Initialization:} $\mathcal{C} =\big\{ \{j\};\ j\in\{1,\dots,p\} \big\}$, $\tilde{\Xb} = \Xb$.
         \WHILE { $|\mathcal{C}| > 1$ }
            \STATE Find a pair of clusters $c_1 ,c_2 \in \mathcal{C}$ which have neighboring voxels and minimize~(\ref{eq:ward_criterion}).
            \STATE $\mathcal{C} \leftarrow \mathcal{C} \setminus \{c_1,c_2\}$.
            \STATE $\mathcal{C} \leftarrow \mathcal{C} \cup (c_1\cup c_2)$.
            \STATE $\tilde{\Xb} \leftarrow [\tilde{\Xb}, \langle \Xb \rangle_{c_1 \cup c_2}]$. 
        \ENDWHILE
   \STATE {\bfseries Return:} $\tilde{\Xb}$.
\end{algorithmic}
\end{algorithm}
Let us now illustrate on the example of linear models (such as those we use in Section~\ref{sec:supervised_framework}) 
what are the implications of considering the augmented space of features.
For a node $j$ of $\mathcal{T}$, we let denote by 
$P_j \subseteq \{1,\dots,p\}$ the set of voxels of the corresponding parcel 
(or, equivalently, the set of leaves of the subtree rooted at node $j$).
In this notation, and for any fMRI volume $\tilde{\xb} \in \Real^q$ in the augmented feature space, 
linear functions indexed by $\wb \in \Real^q$ take the form
$$f_{\wb}(\tilde{\xb})=\wb^\top \tilde{\xb} 
=\sum_{j=1}^q \wb_j \Big [ \frac{1}{|P_j|} \sum_{k \in P_j} \xb_k  \Big ]
=\sum_{k=1}^p \Big [ \sum_{j \in A_k} \frac{\wb_j}{|P_j|} \Big ] \, \xb_k,$$ 
where $A_k$ stands for the set of ancestors of 
a node $k$ in $\mathcal{T}$ (including itself).

\subsection{Hierarchical sparsity-inducing norms} 

In the perspective of inter-subject validation, the augmented space of
variables can be exploited in the following way: Since the information of
single voxels may be unreliable, \textit{the deeper the node in $\mathcal{T}$,
the more variable the corresponding parcel's intensity is likely to be across
subjects}. This property suggests that, while looking for sparse solutions
of~(\ref{eq:opt_pb}), we should preferentially select the variables near the
root of $\mathcal{T}$, before trying to access smaller parcels located further
down in $\mathcal{T}$.

Traditional sparsity-inducing penalties, \emph{e.g.}, the $\ell_1$-norm
$\Omega(\wb)=\sum_{j=1}^p|\wb_j|$, yield sparsity at the level of single
variables $\wb_j$, disregarding potential structures---for instance,
spatial---existing between larger subsets of variables. We leverage here the
concept of 
\textit{structured sparsity}~\cite{Baraniuk2008, Cevher2008, Zhao2009, Huang2009, Jenatton2009, Jacob2009, Micchelli2010, Duarte2011}, 
where $\Omega$ penalizes some
predefined subsets, or \textit{groups}, of variables that reflect prior
information about the problem at hand. 

When these groups form a \textit{partition} of the space of variables, 
the resulting penalty has been shown to
improve the prediction performance and/or interpretability of the learned models, provided that
the block structure is relevant (e.g., see~\cite{Turlach2005,Yuan2006,Liu2009,Kowalski2009,Stojnic2009,Huang2010} and references therein).

If the groups overlap~\cite{Baraniuk2008, Zhao2009, Huang2009, Jenatton2009, Jacob2009, Liu2010d}, 
richer structures can then be encoded. 
In particular, we follow~\cite{Zhao2009} who first introduced
hierarchical sparsity-inducing penalties. Given a node $j$ of $\mathcal{T}$, we
denote by $g_j\subseteq\{1,\dots,q\}$ the set of indices that record all the
descendants of $j$ in $\mathcal{T}$, including itself. In other words, $g_j$
contains the indices of the subtree rooted at $j$; see Figure~\ref{fig:tree}.
If we now denote by $\G$ the set of all $g_j,\ j\in\{1,\dots,q\}$, that is, $\G\defin\{g_1,\dots,g_q\}$, 
we can define our hierarchical penalty as
\begin{equation}\label{eq:omega}
 \Omega(\wb)\defin\sum_{g\in\G}\|\wb_g\|_2\defin\sum_{g\in\G}\Big[ \sum_{j\in g}\wb_j^2 \Big]^{1/2}.
\end{equation}
As formally shown in~\cite{Jenatton2009}, $\Omega$ is a norm on $\Real^q$, and it promotes sparsity at the
level of groups $g \in \G$, in the sense that it acts as a $\ell_1$-norm on the
vector $(\|\wb_g\|_2)_{g\in\G}$. Regularizing by $\Omega$ therefore causes some
$\|\wb_g\|_2$ (and equivalently $\wb_g$) to be zeroed out for some $g \in \G$.
Moreover, since the groups $g \in \G$ represent rooted subtrees of
$\mathcal{T}$, this implies that if one node/parcel $j \in g$ is set to zero by
$\Omega$, the same occurs for all its descendants~\cite{Zhao2009}. To put it
differently, \textit{if one parcel is selected, then all the ancestral parcels
in $\mathcal{T}$ will also be selected}. This property is likely to increase the robustness of the methods to
voxel misalignments between subjects, since large parcels will be considered for addition in the model
before smaller ones.

The family of norms with the previous property is actually slightly larger 
and we consider throughout the paper norms  $\Omega$ of the form~\cite{Zhao2009}:
\begin{equation}\label{eq:omega_weight}
 \Omega(\wb)\defin\sum_{g\in\G} \eta_g \|\wb_g\|,
\end{equation}
where $\|\wb_g\|$ denotes
either the $\ell_2$-norm $\|\wb_g\|_2$ or the $\ell_\infty$-norm
$\|\wb_g\|_\infty\defin\max_{j\in g}|\wb_j|$ and $(\eta_g)_{g\in\G}$ are (strictly) positive weights that can compensate for
the fact that some features are overpenalized as a result of being included in a larger number of groups than others. In light of the results
from~\cite{Jenatton2010b}, we will see in Section~\ref{sec:opt} that a large
class of optimization problems regularized by $\Omega$---as defined
in~(\ref{eq:omega_weight})--- can be solved efficiently.

\section{Supervised learning framework}\label{sec:supervised_framework}

In this section, we introduce the formulations we consider in our experiments.
As further discussed in Section~\ref{sec:exp}, the target $y$ we
try to predict corresponds to (discrete) sizes of objects, \emph{i.e.}, a
one-dimensional \emph{ordered} variable. It is therefore sensible to address
this prediction task from both a regression and a classification viewpoint.
In the remainder of this section, we shall denote by $\{\wb^*,b^*\}$ (or $\{\Wb^*,\bb^*\}$) \textit{a} solution of the optimization problems we present below.\footnote{In the absence of strong convexity,
we cannot in general guarantee the uniqueness of $\wb^*$ (and $\Wb^*$).}
For simplicity, the formulations we review next are all expressed in terms of a matrix $\Xb \in \Real^{n\times p}$ with $p$-dimensional parameters, but they are of course immediately applicable
to the  augmented data $\tilde{\Xb} \in \Real^{n\times q}$ and $q$-dimensional parameters.

\subsection{Regression}\label{sub:regression}

In this first setting, we naturally consider the squared loss function, so that problem~(\ref{eq:opt_pb}) can be reduced to
\begin{equation*}
\min_{\wb \in \Real^p} \frac{1}{2n}\|\yb-\Xb\wb\|^2_2 + \lambda \Omega(\wb)\quad
\text{with}\quad \lambda \geq 0.
\end{equation*}
Note that in this case, we have omitted the intercept $b$ since we can center the vector $\yb$ and the columns of $\Xb$ instead.
Prediction for a new fMRI volume $\xb$ is then simply performed by computing the dot product $\xb^\top\! \wb^*$.

\subsection{Classification}\label{sub:classification}

We can look at our prediction task from a multi-class classification viewpoint.
Specifically, we assume that $\mathcal{Y}$ is a finite set of integers
$\{1,\dots,c\},\ c>2$, and consider both multi-class and ``one-versus-all''
strategies~\cite{Rifkin2004}. We need to slightly extend the
formulation~(\ref{eq:opt_pb}): To this end, we introduce the weight matrix
$\Wb\defin[\wb^1,\dots,\wb^c]\in\Real^{p\times c}$, composed of $c$ weight
vectors, along with a vector of intercepts $\bb \in \Real^c$.

A standard way of addressing multi-class classification problems consists in
using a multi-logit model, also known as multinomial logistic regression~(see,
\emph{e.g.}, \cite{Hastie2009} and references therein). In this case,
class-conditional probabilities are modeled for each class by a softmax
function, namely, given a fMRI volume $\xb$, the probability of having the $k$-th class label reads 
\begin{equation}\label{eq:multiprob}
 \text{Prob}(y=k|\xb; \Wb,\bb) = \frac{\exp\{\xb^\top \wb^k+\bb_k\}}{\sum_{r=1}^c \exp\{\xb^\top \wb^r+\bb_r\}}
 \quad \text{for}\quad k\in \{1,\dots,c\}.
\end{equation}
The parameters $\{\Wb,\bb\}$ are then learned by maximizing the resulting (conditional) log-likelihood, which leads to the following optimization problem:
\begin{equation*}
 \min_{\substack{\Wb\in\Real^{p\times c}\\ \bb \in \Real^c}}
 \frac{1}{n} \sum_{i=1}^n \log\Big[ \sum_{k=1}^c e^{ \xb_i^\top (\wb^k - \wb^{\yb_i}) + \bb_k - \bb_{\yb_i} }
 \Big]
 +\lambda \sum_{k=1}^c \Omega(\wb^k) \enspace .
\end{equation*}
Whereas the regularization term is separable with respect to the different
weight vectors $\wb^k$, the loss function induces a coupling in the columns of
$\Wb$. As a result, the optimization has to be carried out over the entire matrix
$\Wb$.
In this setting, and given a new fMRI volume $\xb$,
we make predictions by choosing the label that maximizes the class-conditional probabilities~(\ref{eq:multiprob}), that is, 
$\text{argmax}_{k\in \{1,\dots,c\}} \text{Prob}(y=k|\xb; \Wb^*,\bb^*)$.

In Section~\ref{sec:exp}, we consider another multi-class classification
scheme. The ``one-versus-all'' strategy (OVA) consists in training $c$ different
(real-valued) binary classifiers, each one being trained to distinguish the examples
in a single class from the observations in all remaining classes. In order to
classify a new example, among the $c$ classifiers, the one which outputs the
largest (most positive) value is chosen. In this framework, we consider binary
classifiers built from both the squared and the logistic loss functions. If we
denote by ${\bar \Yb} \in \{-1,1\}^{n\times c}$ the indicator response matrix
defined as ${\bar \Yb}_i^k\defin1$ if $\yb_i=k$ and $-1$ otherwise, we obtain
\begin{equation*}
 \min_{\substack{\Wb\in\Real^{p\times c}}}
 \frac{1}{2n} \sum_{k=1}^c \|{\bar \Yb}^k - \Xb\wb^k\|_2^2
 +\lambda \sum_{k=1}^c \Omega(\wb^k),
\end{equation*}
and
\begin{equation*}
 \min_{\substack{\Wb\in\Real^{p\times c}\\ \bb \in \Real^c}}
 \frac{1}{n} \sum_{i=1}^n \sum_{k=1}^c \log\Big[ 1 +  e^{ - {\bar \Yb}_i^k( \xb_i^\top \wb^k + \bb_k ) }
 \Big]
 +\lambda \sum_{k=1}^c \Omega(\wb^k).
\end{equation*}
By invoking the same arguments as in Section~\ref{sub:regression},
the vector of intercepts $\bb$ is again omitted in the above problem with the squared loss.
Moreover, given a new fMRI volume $\xb$, we predict the label $k$ that maximizes the response $\xb^\top \! [\wb^*]^k$ among the $c$ different classifiers.
The case of the logistic loss function parallels the setting of the multinomial logistic regression, 
where each of the $c$ ``one-versus-all'' classifiers leads to a class-conditional probability; the predicted label is the one corresponding to the highest probability.

The formulations we have reviewed in this section can be solved efficiently within the same optimization framework
that we now introduce.

\section{Optimization}\label{sec:opt}

The convex minimization problem~(\ref{eq:opt_pb}) is challenging,
since the penalty $\Omega$ as defined in~(\ref{eq:omega_weight}) is non-smooth and the
number of variables to consider is large
(we have $q \approx 10^5$ variables in the following experiments).
These difficulties are well addressed by \emph{forward-backward splitting methods},
which belong to the broader class of proximal methods.
Forward-backward splitting schemes date back (at least) to~\cite{Martinet1970,Lions1979} and
have been further analyzed in various settings (e.g., see~\cite{Tseng1991,Chen1997,Combettes2006}); for a thorough review of proximal splitting techniques, 
we refer the interested readers to~\cite{Combettes2010}.

Our convex minimization problem~(\ref{eq:opt_pb}) can be handled well by such techniques 
since it is the sum of two semi-lower continuous, proper, convex functions with non-empty domain, 
and where one element---the loss function $\mathcal{L}(\yb,\Xb,.)$---is assumed differentiable with Lipschitz-continuous gradient
(which notably covers the cases of the squared and simple/multinomial logistic functions, as introduced in Section~\ref{sec:supervised_framework}).

To describe the principle of forward-backward splitting methods, we need to introduce the concept of \textit{proximal operator}. 
The proximal operator associated with our regularization term $\lambda\Omega$, which we denote by $\text{Prox}_{\lambda \Omega}$, 
is the function that maps a vector $\wb \in \Real^p$ to the unique solution of
\begin{equation}\label{eq:pb_prox}
   \min_{\vb \in \Real^p} \frac{1}{2} \| \vb - \wb  \|_2^2 + \lambda  \Omega(\vb).
\end{equation}
This operator was initially introduced by Moreau~\cite{Moreau1962} to generalize the 
projection operator onto a convex set; for a complete study of the properties of $\text{Prox}_{\lambda \Omega}$, see~\cite{Combettes2006}.
Based on definition~(\ref{eq:pb_prox}), and given the current iterate $\wb^{(k)}$,\footnote{For clarity of the presentation,
we do not consider the optimization of the intercept that we leave unregularized in all our experiments.} 
the typical update rule of forward-backward splitting methods has the form\footnote{For simplicity, we only present a constant-stepsize scheme; 
adaptive line search can also be used in this context and can lead to larger stepsizes~\cite{Combettes2010}.}
\begin{equation}\label{eq:update_ista}
\wb^{(k+1)} \leftarrow \text{Prox}_{\frac{\lambda}{L}\Omega}\Big(  \wb^{(k)} - \frac{1}{L} \nabla \mathcal{L}_\wb\!(\yb,\Xb,\wb^{(k)})  \Big),
\end{equation}
where $L>0$ is a parameter which is a upper bound on the Lipschitz constant of the gradient of $\mathcal{L}$.
In the light of the update rule~(\ref{eq:update_ista}), 
we can see that solving efficiently problem~(\ref{eq:pb_prox}) is crucial to enjoy good performance. 
In addition, when the non-smooth term $\Omega$ is not present, the previous proximal problem~(\ref{eq:update_ista}),
also known as the implicit or backward step,
exactly leads to the standard gradient update rule.

For many regularizations $\Omega$ of interest, the solution of problem~(\ref{eq:pb_prox}) can actually be computed in closed form in simple settings: in particular, when $\Omega$ is the $\ell_1$-norm, the proximal operator is the well-known soft-thresholding operator~\cite{Donoho1995}.
The work of \cite{Jenatton2010b}
recently showed that the proximal problem~(\ref{eq:pb_prox}) could be solved efficiently and exactly with
$\Omega$ as defined in (\ref{eq:omega_weight}).
The underlying idea of this computation is to solve a \emph{well-ordered} sequence of
simple proximal problems associated with each of the terms $\|\wb_g\|$ for $g\in\G$.
We refer the interested readers to~\cite{Jenatton2010b} for further details on this norm and to~\cite{Bach2010} for a broader view.

In the subsequent experiments, 
we focus on accelerated multi-step versions of forward-backward splitting methods (see, e.g., \cite{Nesterov2007, Beck2009, Wright2009}),\footnote{More precisely, 
we use the accelerated proximal gradient scheme (FISTA) taken from~\cite{Beck2009}. 
The \texttt{Matlab/C++} implementation we use 
is available at \texttt{http://www.di.ens.fr/willow/SPAMS/}.}
where the proximal problem~(\ref{eq:update_ista}) is not solved for a current estimate, but for an auxiliary sequence
of points that are linear combinations of past estimates.
These accelerated versions have increasingly drawn the attention of a broad research community since
they can deal with large non-smooth convex problems, and
their convergence rates on the objective achieve the complexity bound of $O(1/k^2)$, with $k$ denoting the iteration number. 
As a side comment, note that as opposed to standard one-step forward-backward splitting methods, nothing can be said about the convergence of the sequence of iterates themselves.
In our case, the cost of each iteration is dominated by the computation of the gradient 
(\emph{e.g.}, $O(np)$ for the squared loss) and the proximal operator, whose time complexity is linear, or close to linear, 
in $p$ for the tree-structured regularization~\cite{Jenatton2010b}.
\section{Experiments and results}\label{sec:exp}

We now present experimental results on simulated data and real fMRI data.

\subsection{Simulations}\label{sec:simu}

\begin{figure}[tb]
    \centering
        \includegraphics[width=\linewidth]{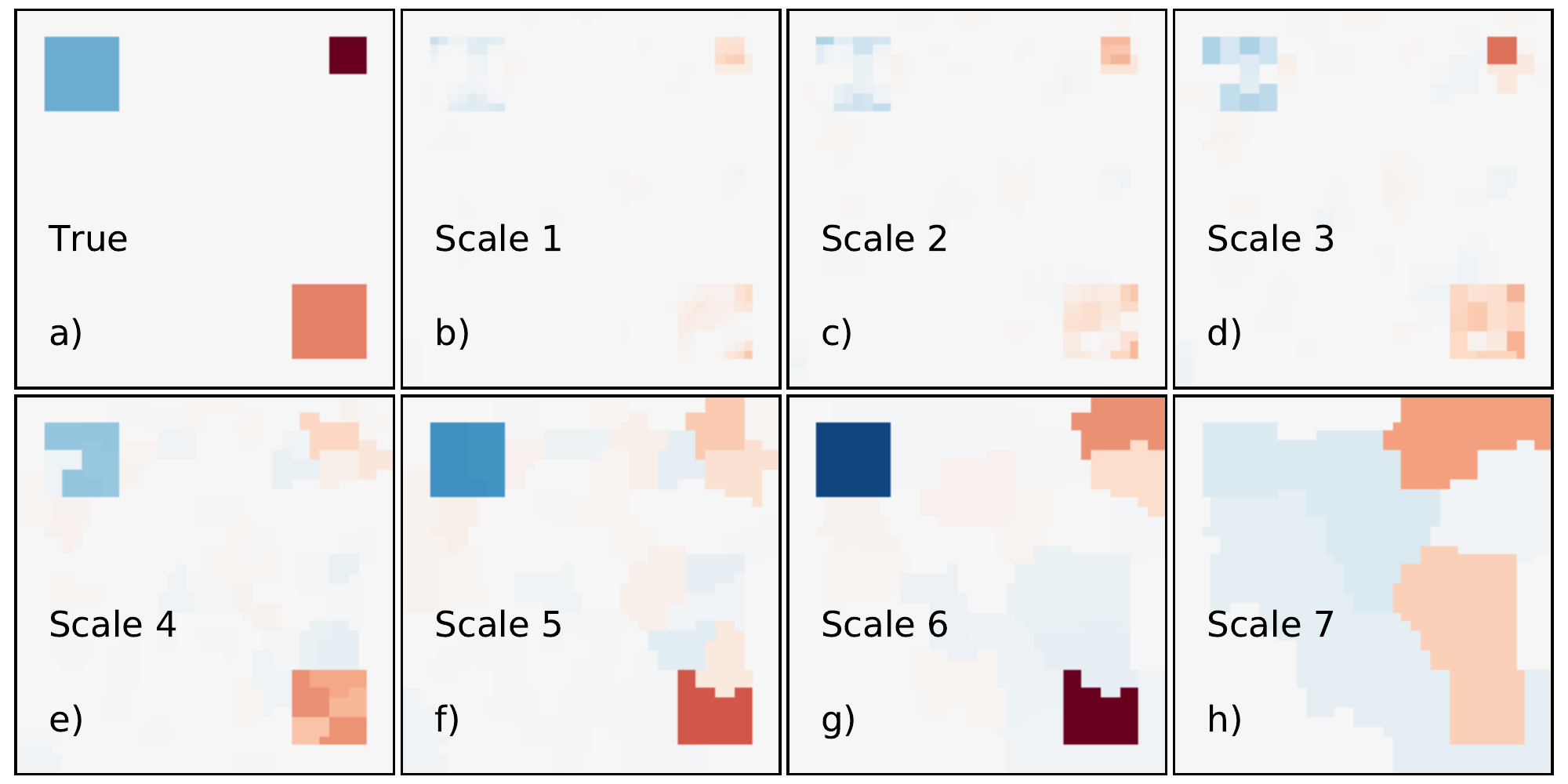}
    \caption{Weights $\wb^*$ estimated in the simulation study. The true coefficients
    are presented in a) and the estimated weights at different scales,
    \emph{i.e.,} different depths in the tree, are presented in b)-h)
    with the same colormap. The results are best seen in color.}
    \label{fig:tree_weights}
\end{figure}

In order to illustrate the proposed method, the hierarchical regularization
with the $\ell_2$-norm and $\eta_g=1$ for all $g\in\G$ was applied in a regression
setting on a small two-dimensional simulated dataset consisting of 300
square images ($40 \times 40$ pixels i.e. $\Xb \in \Real^{300 \times 1600}$).
The weight vector $\wb$
used in the simulation--- itself an image of the same dimension--- is presented in Fig.~\ref{fig:tree_weights}-a. It
consists of three localized regions of two different sizes that are predictive of the output.
The images $\xb^{(i)}$ are sampled so as to obtain a correlation structure which mimics fMRI data. Precisely, each image $\xb^{(i)}$ was obtained by smoothing
a completely random image --- where each pixel was drawn i.i.d from a normal distribution --- with a Gaussian kernel (standard deviation 2 pixels), which introduces spatial correlations between
neighboring pixels. Subsequently, additional correlations between the regions corresponding to the
three patterns were introduced
in order to simulate co-activations between different brain regions, by multiplying the signal by the square-root of an appropriate covariance matrix $\Sigmab$.
Specifically, $\Sigmab \in \Real^{1600 \times 1600}$ is a spatial covariance between voxels, with diagonal set to $\Sigmab_{i,i}=1$ for all $i$,
and with two off-diagonal blocks. Let us denote $\mathcal{C}_1$ and $\mathcal{C}_2$
the set of voxels forming the two larger patterns, and $\mathcal{C}_3$ the voxels
in the small pattern. The covariance coefficients are set to $\Sigmab_{i,j}=0.3$ for
$i \in \mathcal{C}_1$ and $j \in \mathcal{C}_2$, and $\Sigmab_{i,j}=-0.2$
for $i \in \mathcal{C}_2$ and $j \in \mathcal{C}_3$. The covariance
is of course symmetric.

The choice of the weights and of the correlation introduced in images aim at
illustrating how the hierarchical regularization estimates weights at different
resolutions in the image. The targets were simulated
by forming $\wb ^\top \xb^{(i)}$ corrupted with an additive white noise (SNR=10dB). 
The loss used was the squared loss as detailed in Section~\ref{sub:regression}. The regularization parameter
was estimated with two-fold cross-validation (150 images per fold) on a
logarithmic grid of 30 values between $10^3$ and $10^{-3}$.

The components of the estimated weight vector $\wb^*$ at different scales are presented in the images of Fig.~\ref{fig:tree_weights} , 
with each image corresponding to a different depth in the tree.
For a given tree depth, an image is formed from the corresponding
parcellation. All the voxels within a parcel are colored
according to the associated scalar in $\wb^*$.
It can be observed that all three patterns are present in the weight
vector but at different depths in the tree. The small activation
in the top-right hand corner shows up mainly at scale 3 while
the bigger patterns appear higher in the tree at scales 5 and 6.
This simulation clearly illustrates the ability of the method
to capture informative spatial patterns at different scales.

\subsection{Description of the fMRI data}

We apply the different methods to analyze the data of ten subjects from an fMRI
study originally designed to investigate object coding in visual cortex 
(see \cite{eger2007} for details).
During the experiment, ten healthy volunteers viewed objects of two categories
(each one of the two categories is used in half of the subjects) with four
different exemplars in each category. Each exemplar was presented at three
different sizes (yielding $12$ different experimental conditions per subject).
Each stimulus was presented four times in each of the six sessions. We averaged
data from the four repetitions, resulting in a total of $n=72$ volumes by
subject (one volume of each stimulus by session).
Functional volumes were acquired on a 3-T MR system with eight-channel
head coil (Siemens Trio, Erlangen, Germany) as T2*-weighted
echo-planar image (EPI) volumes. Twenty transverse slices were
obtained with a repetition time of 2s (echo time, 30ms; flip angle,
$70^{\circ}$; $2\times2\times2$-mm voxels; $0.5$-mm gap).
Realignment, spatial normalization to MNI space, slice timing correction and GLM
fit were performed with the SPM5 software\footnote{\texttt{http://www.fil.ion.ucl.ac.uk/spm/software/spm5}.}.
In the GLM, the time course of each of the $12$ stimuli convolved with a
standard hemodynamic response function was modeled separately, while accounting
for serial auto-correlation with an AR(1) model and removing low-frequency
drift terms with a high-pass filter with a cut-off of 128s ($7.8\times10^{-3}$Hz).
In the present work we used the resulting session-wise parameter estimate volumes.
Contrary to a common practice in the field the data were not smoothed with
an isotropic Gaussian filter. All the analysis are performed 
on the whole acquired volume.

The four different exemplars in each of the two categories were pooled, leading
to volumes labeled according to the three possible sizes of the object. By doing so,
we are interested in finding discriminative information to predict the size of
the presented object.

This can be reduced to either a regression problem in which our goal
is to predict the class label of the size of the presented object (i.e., $y \in \{0,1,2\}$),\footnote{An interesting alternative 
would be to consider some real-valued dimension such as the field of view of the object.}
or a three-category classification problem, each size corresponding to a category.
We perform an inter-subject analysis on the sizes both in regression and classification settings. This analysis
relies on subject-specific fixed-effects activations, \emph{i.e.}, for each
condition, the six activation maps corresponding to the six sessions are
averaged together. This yields a total of 12 volumes per subject, one for
each experimental condition. The dimensions of the real data set are 
$p \approx 7\times 10^4$ and $n=120$ (divided into three different sizes).
We evaluate the performance of the method by
cross-validation with a natural data splitting, \emph{leave-one-subject-out}.
Each fold consists of 12 volumes.
The parameter $\lambda$ of all methods is optimized over a grid of 30 values of the form $2^k$,
with a nested leave-one-subject-out cross-validation on the training set.
The exact scaling of the grid varies for each model to account for different $\Omega$.

\subsection{Methods involved in the comparisons}

In addition to considering standard $\ell_1$- and squared $\ell_2$-regularizations
in both our regression and multi-class classification tasks,
we compare various methods that we now review.

First of all, when the regularization $\Omega$ as defined in~(\ref{eq:omega_weight}) is employed,
we consider three settings of values for $(\eta_g)_{g\in\G}$ which leverage the tree structure $\mathcal{T}$.
More precisely, we set $\eta_g=\rho^{\textrm{depth}(g)}$ for $g$ in $\G$, with $\rho \in \{0.5, 1, 1.5\}$ and
where $\textrm{depth}(g)$ denotes the depth of the root of the group $g$ in $\mathcal{T}$.
In other words, the larger $\rho$, the more averse we are to selecting small (and variable) parcels located near the leaves of $\mathcal{T}$.
As the results illustrate it, the choice of $\rho$ can have a significant impact on the performance.
More generally, the problem of selecting $\rho$ properly is a difficult question which is still under investigation, both theoretically and practically, e.g., see~\cite{Bach2009}.

The greedy approach from~\cite{michel2010} is included in the comparisons,
for both the regression and classification tasks.
It relies on a top-down exploration of the tree $\mathcal{T}$. 
In short, starting from the root parcel that contains all the voxels, 
we choose at each step the split of the parcel that yields the highest prediction score.  
The exploration step is performed until a given number of parcels is reached, 
and yields a set of nested parcellations with increasing complexity. 
Similarly to a model selection step, we chose the best parcellation among those found in the exploration step. 
The selected parcellation is thus used on the test set.
In the regression setting, this approach is combined with Bayesian ridge regression, 
while it is associated with a linear support vector machine for the classification task 
(whose regularization parameter, often referred to as $C$ in the literature~\cite{Scholkopf2002}, is found by nested cross-validation in $\{0.01, 0.1, 1\}$).

\subsubsection{Regression setting}

In order to evaluate whether the level of sparsity is critical in our analysis,
we implemented a reweighted $\ell_1$-scheme~\cite{Candes2008}.
In this case, sparsity is encouraged more aggressively as a multi-stage convex relaxation of a concave penalty.
Specifically, it consists in using iteratively a weighted $\ell_1$-norm,
whose weights are determined by the solution of previous iteration.
Moreover, we additionally compare to Elastic net~\cite{zou2005}, whose second regularization parameter 
is set by cross-validation as a fraction of $\lambda$, that is, $ \alpha \lambda$ 
with $\alpha \in \{0.5, 0.05, 0.005, 0.0005\}$.

To better understand the added value of the hierarchical norm~(\ref{eq:omega_weight}) over unstructured penalties,
we not only consider the plain $\ell_1$-norm in the augmented feature space, 
but also another variant of weighted $\ell_1$-norm.
The weights are manually set and reflect the underlying tree structure $\mathcal{T}$.
By analogy with the choice of $(\eta_g)_{g\in\G}$ made for the tree-structured regularization, 
we take exponential weights depending on the depth of the variable $j$, 
where $\eta_j = \rho^{\textrm{depth}(j)}$ with $\rho \in \{0.5, 1.5\}$.\footnote{Formally,
the depth of the feature $j$ is equal to $\textrm{depth}(g_j)$, 
where $g_j$ is the smallest group in $\G$ that contains $j$ 
(\emph{smallest} is understood here in the sense of the inclusion).} 
We also tried weights $(\eta_j)_{j\in\{1,\dots,p\}}$ that are linear with respect to the depths, i.e.,
$\eta_j=\frac{\textrm{depth}(j)}{\max_{k\in\{1,\dots,p\}}\textrm{depth}(k)}$,
but those led to worse results. In Table~\ref{Tab:res_sizes_inter_reg}, 
we only present the best result of this weighted $\ell_1$-norm, obtained with the exponential weight and $\rho=1.5$.
We now turn to the models taking part in the classification task.

\subsubsection{Classification setting}

As discussed in Section~\ref{sub:classification},
the optimization in the classification setting is carried out over a matrix of weights $\Wb \in \Real^{p \times c}$.
This makes it possible to consider regularization schemes that couple the selection of variables across rows of that matrix.

In particular, we apply ideas from \emph{multi-task}
learning~\cite{Obozinski2009} by viewing each class as a task. More precisely,
we use a regularization norm defined by $\Omega_{\text{multi-task}}(\Wb)\defin
\sum_{j=1}^p \|\Wb_j\|$, where $\|\Wb_j\|$ denotes either the $\ell_2$- or
$\ell_\infty$-norm of the $j$-th row of $\Wb$. The rationale for the definition
of $\Omega_{\text{multi-task}}$ is to assume that the set of relevant voxels is
the same across the $c$ different classes, so that sparsity is induced
simultaneously over the columns of $\Wb$. It should be noted that , in the ``one-versus-all''
setting, although the loss functions for the $c$ classes are decoupled, the use
of $\Omega_{\text{multi-task}}$ induces a relationship that ties the optimization problems together.

Note that the tree-structured regularization $\Omega$ we consider
does not impose a joint pattern-selection across the $c$ different classes.
It would however be possible to use $\Omega$ over the matrix $\Wb$, in a multi-task setting.
More precisely, we would define $\Omega(\Wb) = \sum_{g\in\G} \| \Wb_g \|$, 
where $\| \Wb_g \|$ denotes either the $\ell_2$- or
$\ell_\infty$-norm of the vectorized sub-matrix $\Wb_g \defin [\Wb_{jk}]_{j\in g, k\in\{1,\dots,c\}}$.
This definition constitutes a direct extension of the standard non-overlapping $\ell_1/\ell_2$- and 
$\ell_1/\ell_\infty$-norms used for multi-task.
Furthermore, it is worth noting that the optimization tools from~\cite{Jenatton2010b} would still apply for this tree-structured matrix regularization.

\subsection{Results}

We present results comparing our approach based on the hierarchical sparsity-inducing
norm~(\ref{eq:omega_weight}) with the regularization listed in the
previous section. For each method, we computed the cross-validated prediction
accuracy and the percentage of non-zero coefficients, \emph{i.e.}, the level of
sparsity of the model.

\subsubsection{Regression results}

The results for the inter-subject regression analysis are reported
in Table~\ref{Tab:res_sizes_inter_reg}.
The lowest error in prediction accuracy is obtained by both
the greedy strategy and
the proposed hierarchical structured sparsity approach (Tree $\ell_2$ with $\rho=1$)
whose performances are essentially indistinguishable.
Both also yield one of the lowest standard deviation indicating that the results are
most stable. This can be explained by the fact that 
the use of local signal averages in the proposed algorithm is a good way to get some robustness to inter-subject variability. 
We also notice that the
sparsity-inducing approaches (Lasso and reweighted $\ell_1$) have the highest
error in prediction accuracy, probably because the obtained solutions are
too sparse, and suffer from the absence of perfect voxel-to-voxel
correspondences between subjects.

\begin{figure}[h!tb]
    \begin{minipage}{0.05\linewidth}
        a)
    \end{minipage}%
    \begin{minipage}{0.94\linewidth}
        \begin{center}
         \includegraphics[width=1.\linewidth]
        {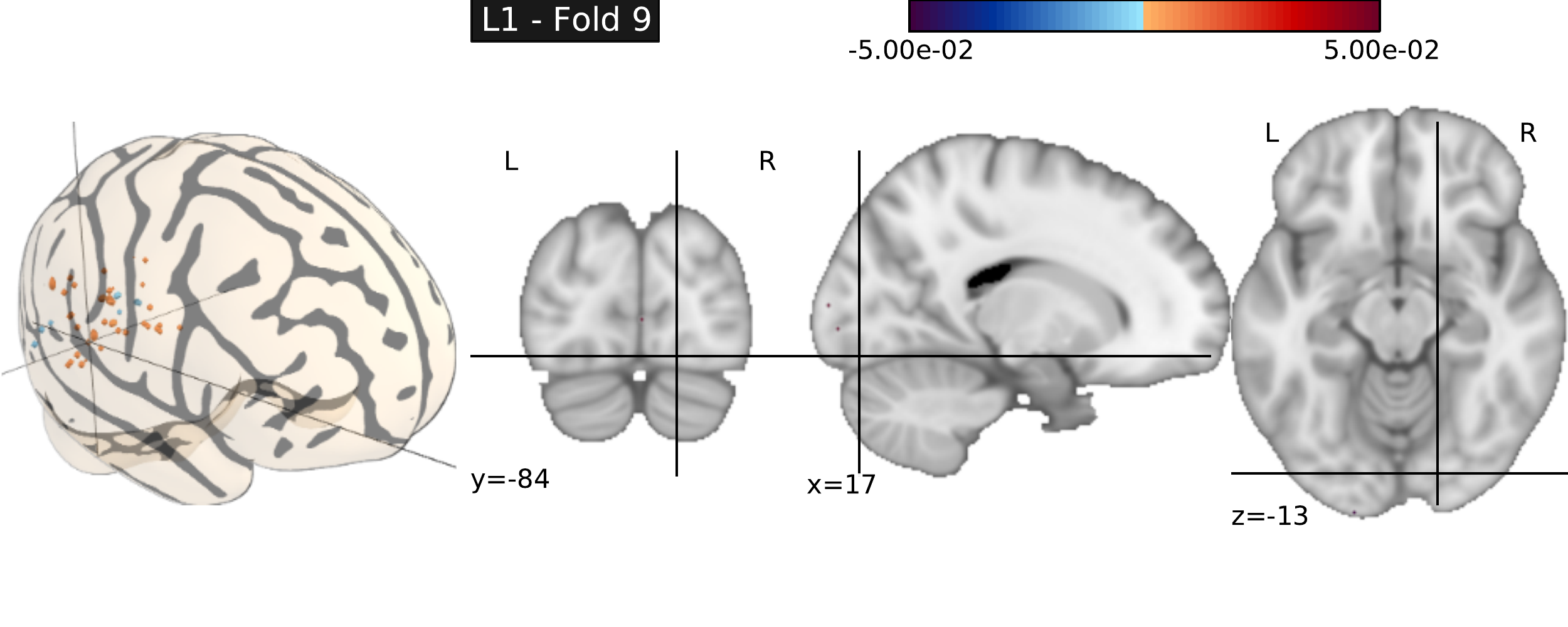}
        \end{center}
    \end{minipage}\\
    \begin{minipage}{0.05\linewidth}
        b)
    \end{minipage}%
    \begin{minipage}{0.94\linewidth}
        \begin{center}
         \includegraphics[width=1.\linewidth]
        {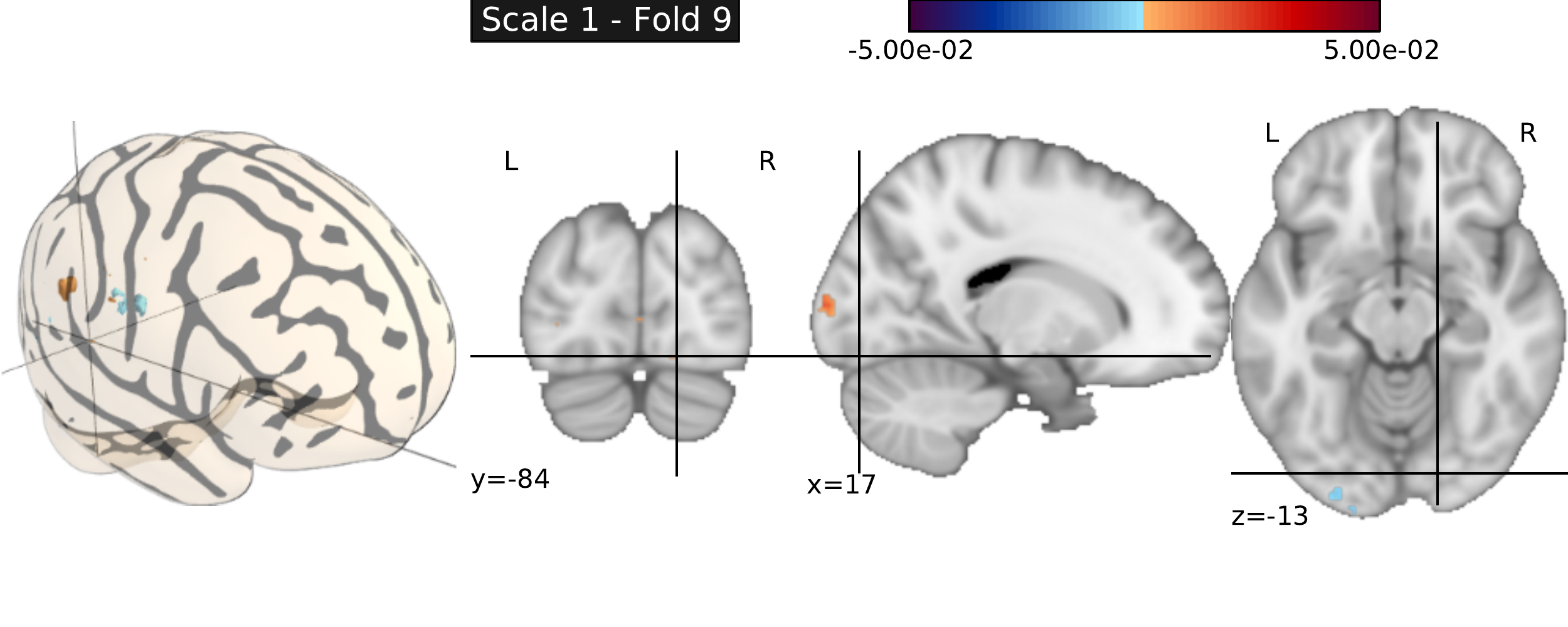}
        \end{center}
    \end{minipage}\\
    \begin{minipage}{0.05\linewidth}
        c)
    \end{minipage}%
    \begin{minipage}{0.94\linewidth}
        \begin{center}
         \includegraphics[width=1.\linewidth]
        {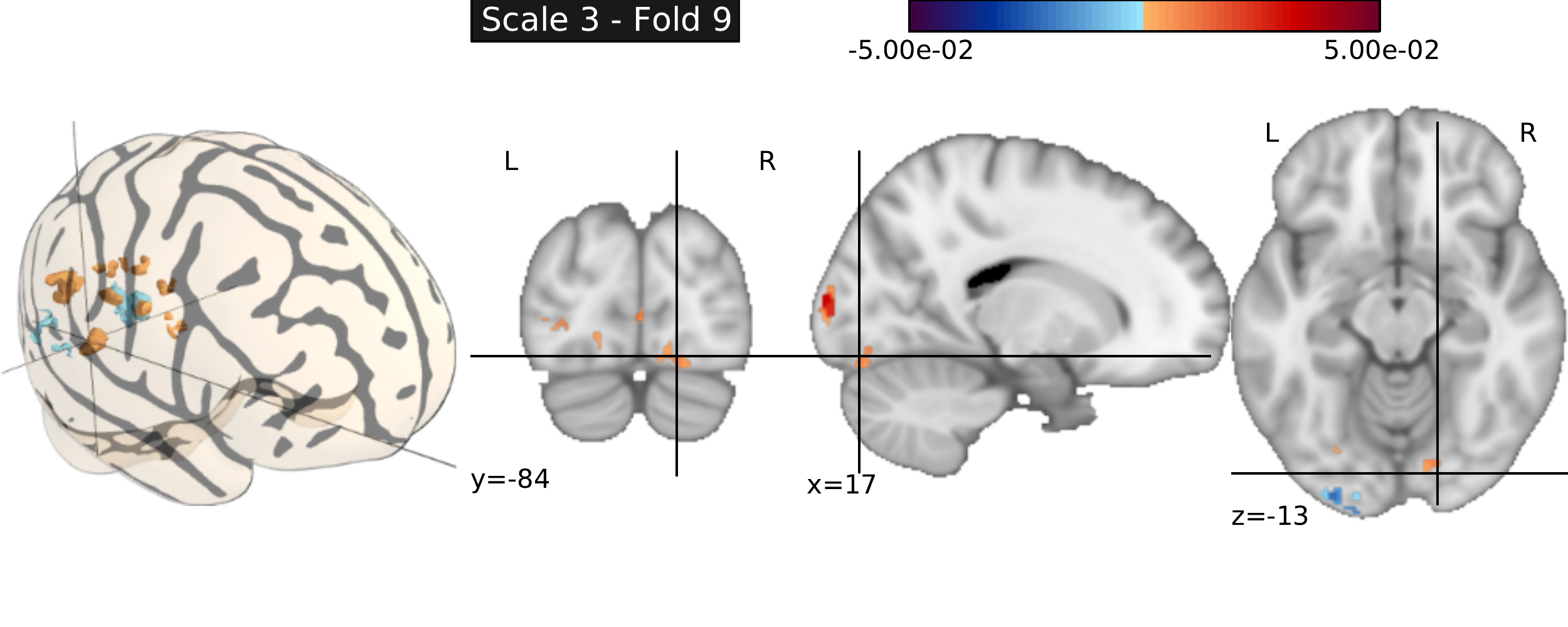}
        \end{center}
    \end{minipage}\\
    \begin{minipage}{0.05\linewidth}
        d)
    \end{minipage}%
    \begin{minipage}{0.94\linewidth}
        \begin{center}
         \includegraphics[width=1.\linewidth]
        {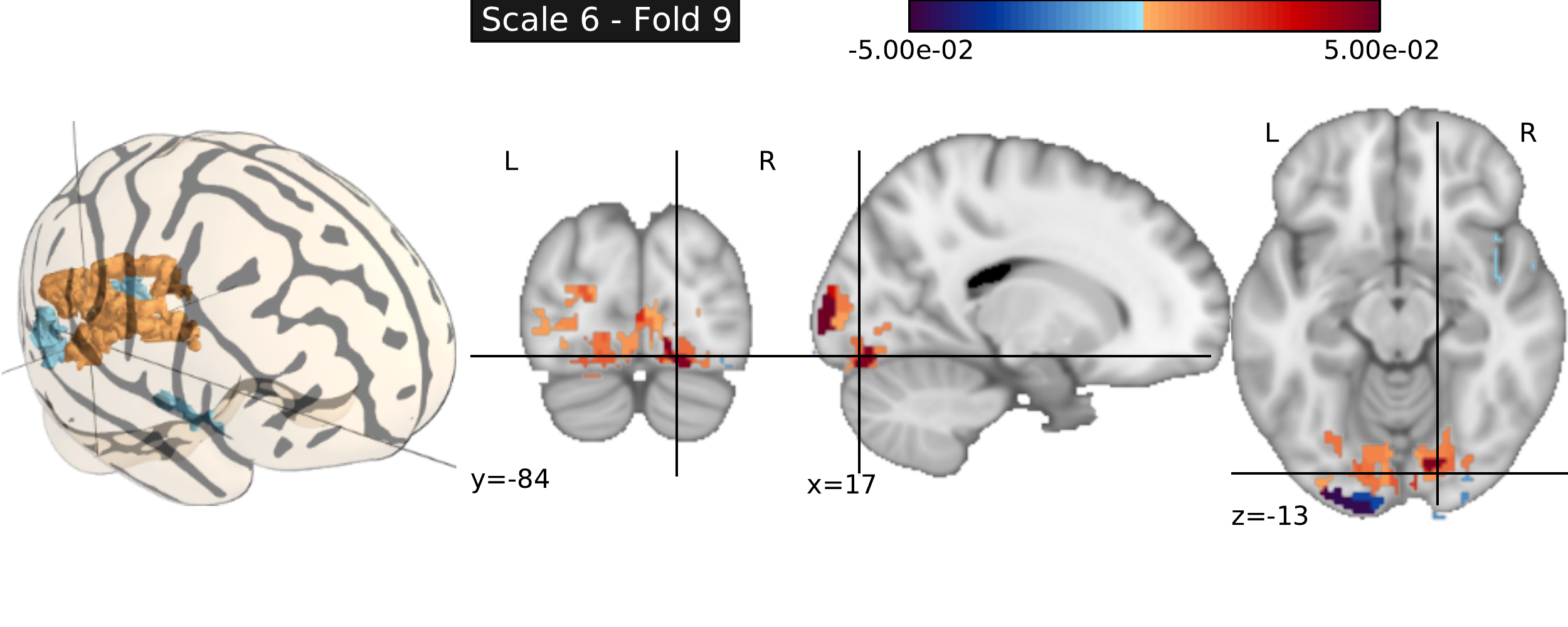}
        \end{center}
    \end{minipage}%
\caption{Maps of weights obtained using different
regularizations in the regression setting.
(a) $\ell_1$ regularization - We can notice that the predictive pattern
obtained is excessively sparse, and is not easily readable despite being mainly
located in the occipital cortex.
(b-d) tree $\ell_2$ regularization ($\rho=1$) at different scales -
In this case, the
regularization algorithm extracts a pattern of voxels with a compact structure,
that clearly outlines 
early visual cortex which is expected to discriminate between stimuli of different sizes. 
3D images were generated with
Mayavi~\cite{ramachandran-varoquaux:10}.}
\label{Fig:fig_w_reg}
\end{figure}

In terms of sparsity, we can see, as expected, that ridge regression does not yield any
sparsity and that the Lasso solution is very sparse (in the feature space, with approximately
$7\times 10^4$ voxels). Our method yields a median value of 9.36\% of non-zero
coefficients (in the augmented space of features, with about $1.4\times 10^5$
nodes in the tree). The maps of weights obtained with Lasso and the hierarchical
regularization for one fold, are reported in Fig.~\ref{Fig:fig_w_reg}. The Lasso yields
a scattered and overly sparse pattern of voxels, that is not easily readable,
while our approach extracts a pattern of voxels with a compact structure, that
clearly outlines brain regions expected to activate differentially for stimuli with different low-level visual properties, 
\emph{e.g.}, sizes;
the early visual cortex in the occipital lobe at the back of the
brain. Interestingly, the patterns of voxels show some symmetry
between left and right hemispheres, especially in the primary
visual cortex which is located at the back and center of the brain.
This observation is consistent with the current understanding in neuroscience
that the symmetric parts of this brain region process respectively the visual contents
of each of the visual hemifields. The weights obtained at different depth level
in the tree, corresponding to different
scales, show that the largest coefficients are concentrated at the higher scales (scale
6 in Fig.~\ref{Fig:fig_w_reg}), which suggest that the object sizes cannot be well
decoded at the voxel level but require features corresponding to larger
clusters of voxels.

{%
\newcommand{\mc}[3]{\multicolumn{#1}{#2}{#3}}
\definecolor{tcA}{rgb}{0.627451,0.627451,0.643137}
\begin{table}
\begin{center}\footnotesize{
\begin{tabular}{|l|lll}\cline{1-4}
Loss function: & \mc{3}{c|}{Squared loss}\\\cline{1-4}
\mc{4}{>{\columncolor{tcA}}l}{×}\\\hline
\mc{1}{>{\columncolor{tcA}}l}{×}
& \mc{1}{c|}{Error (mean,std)}
& \mc{1}{c|}{P-value w.r.t. Tree $\ell_2$ ($\rho=1$)}
& \mc{1}{c|}{Median fraction of non-zeros (\%)}\\\hline
Regularization: & \mc{1}{c|}{×} & \mc{1}{c|}{×} & \mc{1}{c|}{}\\\hline
$\ell_2$ (Ridge) & \mc{1}{c|}{(13.8,\ 7.6)} & \mc{1}{c|}{\!\!\!0.096} & \mc{1}{c|}{100.00}\\\hline
$\ell_1$  & \mc{1}{c|}{(20.2,\ 10.9)} & \mc{1}{c|}{0.013$^*$} & \mc{1}{c|}{0.11}\\\hline
$\ell_1+\ell_2$ (Elastic net)  & \mc{1}{c|}{(14.4,\ 8.8)} & \mc{1}{c|}{\!\!\!0.065} & \mc{1}{c|}{0.14}\\\hline
Reweighted $\ell_1$ & \mc{1}{c|}{(18.8,\ 14.6)} & \mc{1}{c|}{\!\!\!0.052} & \mc{1}{c|}{0.10}\\\hline
$\ell_1$ (augmented space) & \mc{1}{c|}{(14.2,\ 7.9)} & \mc{1}{c|}{\!\!\!0.096} & \mc{1}{c|}{0.02}\\\hline
$\ell_1$ (tree weights) & \mc{1}{c|}{(13.9,\ 7.9)} & \mc{1}{c|}{0.032$^*$} & \mc{1}{c|}{0.02}\\\hline
Tree $\ell_2$ ($\rho=0.5$) & \mc{1}{c|}{(13.0,\ 7.4)} & \mc{1}{c|}{\!\!\!0.137} & \mc{1}{c|}{99.99}\\\hline
Tree $\ell_2$ ($\rho=1$) & \mc{1}{c|}{(\textbf{11.8},\ \textbf{6.7})} & \mc{1}{c|}{-} & \mc{1}{c|}{9.36}\\\hline
Tree $\ell_2$ ($\rho=1.5$)  & \mc{1}{c|}{(13.5,\ 7.0)} & \mc{1}{c|}{\!\!\!0.080} & \mc{1}{c|}{0.04}\\\hline
Tree $\ell_\infty$ ($\rho=0.5$) & \mc{1}{c|}{(13.6,\ 7.8)} & \mc{1}{c|}{\!\!\!0.080} & \mc{1}{c|}{99.99}\\\hline
Tree $\ell_\infty$ ($\rho=1$) & \mc{1}{c|}{(12.8,\ 6.7)} & \mc{1}{c|}{\!\!\!0.137} & \mc{1}{c|}{1.22}\\\hline
Tree $\ell_\infty$ ($\rho=1.5$) & \mc{1}{c|}{(13.0,\ 6.8)} & \mc{1}{c|}{\!\!\!0.096} & \mc{1}{c|}{0.04}\\\hline
\mc{4}{>{\columncolor{tcA}}l}{×}\\\hline
\mc{1}{>{\columncolor{tcA}}l}{×}
& \mc{1}{c|}{Error (mean,std)}
& \mc{1}{c|}{P-value w.r.t. Tree $\ell_2$ ($\rho=1$)}
& \mc{1}{c|}{Median fraction of non-zeros (\%)}\\\hline
Greedy & \mc{1}{c|}{(12.0,\ 5.5)} & \mc{1}{c|}{\!\!\!0.5} & \mc{1}{c|}{0.01}\\\hline
\end{tabular}}
\end{center}
\caption{Prediction results obtained on fMRI data (see text) for the regression
setting. From the left, the first column contains the mean and standard
deviation of the test error (mean squared error), computed over
leave-one-subject-out folds. 
The best performance is obtained with the greedy technique and the hierarchical $\ell_2$ penalization  ($\rho=1$) constructed from the Ward tree.
Methods with performance significantly worse than the latter is assessed by Wilcoxon two-sample paired
signed rank tests (The superscript $^*$ indicates a rejection at $5\%$). Levels of sparsity reported are in the augmented space whenever it is used.}
\label{Tab:res_sizes_inter_reg}
\end{table}
}%

\subsubsection{Classification results}

The results for the inter-subject classification analysis are reported
in Table~\ref{Tab:res_sizes_inter_cl}. The best performance is
obtained with a multinomial logistic loss function, also
using the hierarchical $\ell_2$ penalization  ($\rho=1$).

It should be noted that the sparsity level of the different model estimated vary widely depending
on the loss and regularization used. With the squared loss, $\ell_1$ type regularization, including
the multi-task regularizations based on the $\ell_1/\ell_2$ and $\ell_1/\ell_\infty$ norm tend to select
quite sparse models, which keep around $0.1\%$ of the voxels. When using logistic type losses, these regularizations
tend to select a significantly large number of voxels, which suggests that the selection problem is really difficult and 
that these methods are unstable. For the methods with hierarchical regularization, on the contrary, the sparsity tends to improve with the choice
of loss functions that are better suited to the classification problem and tuning $\rho$ trades off smoothly sparsity of the model against performance, 
from models that are not sparse when $\rho$ is small to very sparse models when $\rho$ is large. 
In particular a better compromise between sparsity and prediction performance can probably be obtained by tuning $\rho \in [1,1.5]$.

\begin{figure}[h!tb]
    \begin{minipage}{0.05\linewidth}
        a)
    \end{minipage}%
    \begin{minipage}{0.94\linewidth}
        \begin{center}
         \includegraphics[width=1.\linewidth]
        {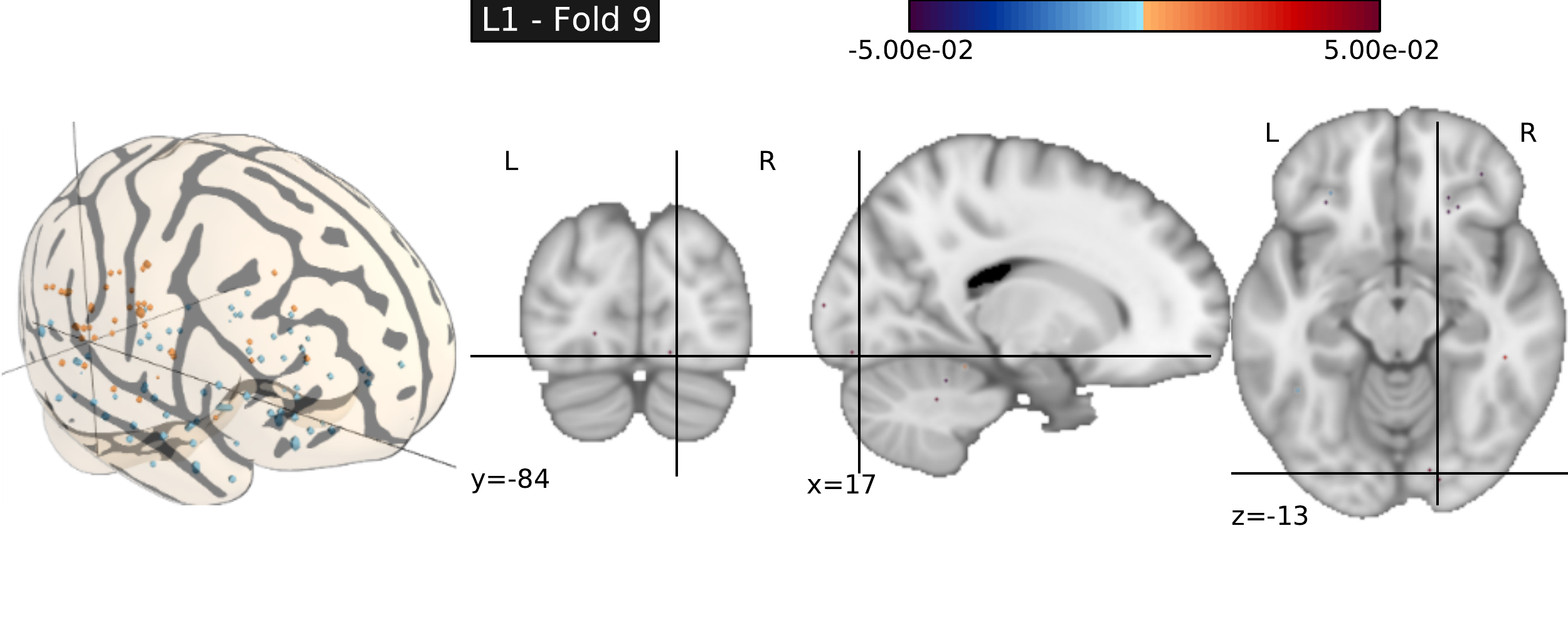}
        \end{center}
    \end{minipage}\\
    \begin{minipage}{0.05\linewidth}
        b)
    \end{minipage}%
    \begin{minipage}{0.94\linewidth}
        \begin{center}
         \includegraphics[width=1.\linewidth]
        {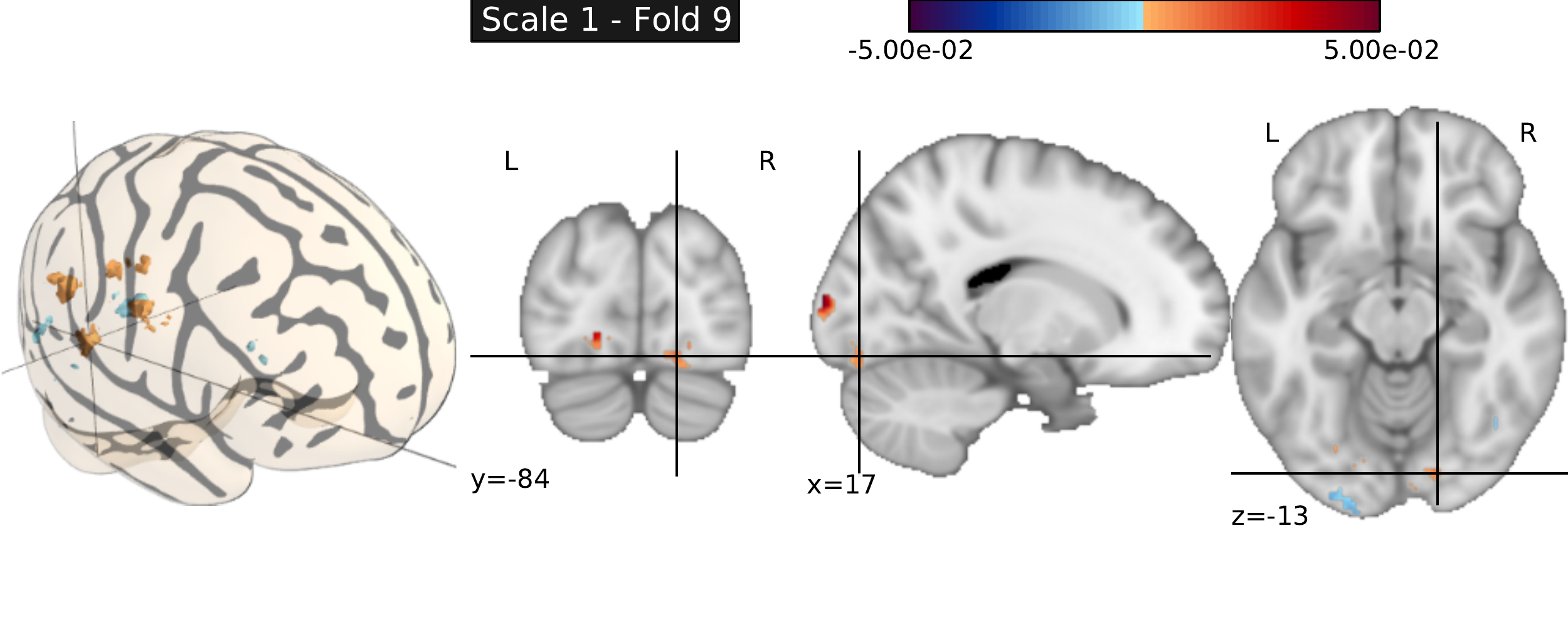}
        \end{center}
    \end{minipage}\\
    \begin{minipage}{0.05\linewidth}
        c)
    \end{minipage}%
    \begin{minipage}{0.94\linewidth}
        \begin{center}
         \includegraphics[width=1.\linewidth]
        {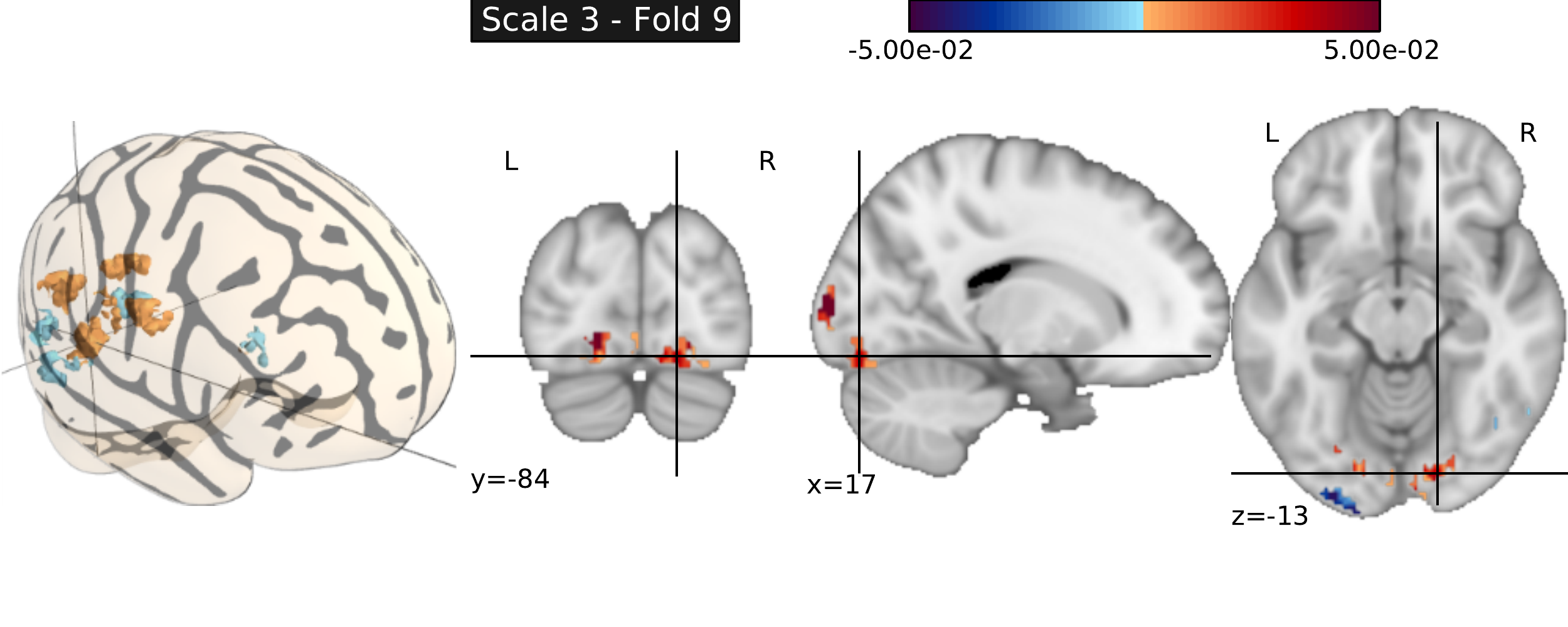}
        \end{center}
    \end{minipage}\\
    \begin{minipage}{0.05\linewidth}
        d)
    \end{minipage}%
    \begin{minipage}{0.94\linewidth}
        \begin{center}
         \includegraphics[width=1.\linewidth]
        {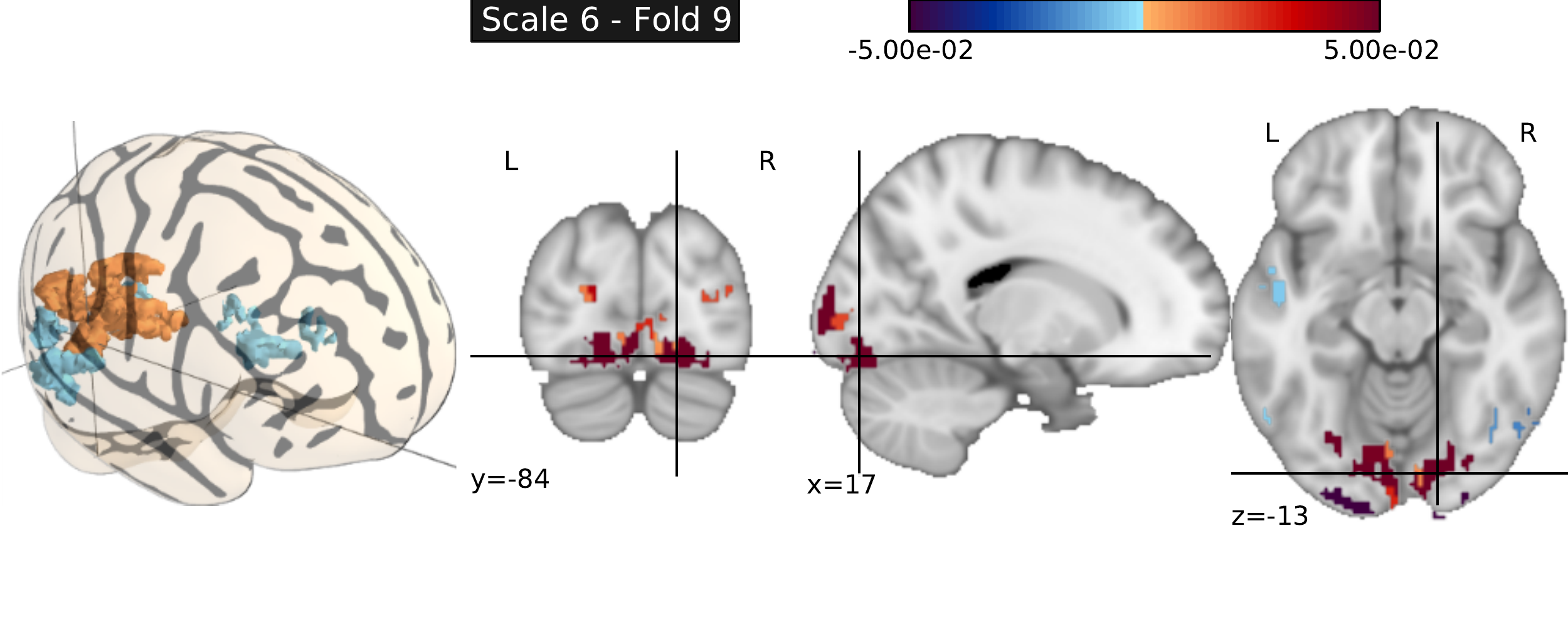}
        \end{center}
    \end{minipage}%
\caption{Maps of weights obtained using different
regularizations in the classification setting.
(a) $\ell_1$ regularization (with squared loss) - We can notice that the predictive pattern
obtained is excessively sparse, and is not easily readable with voxels scattered all over the brain.
(b-d) tree regularization (with multinomial logistic loss) at different scales - In this case, the
regularization algorithm extracts a pattern of voxels with a compact structure,
that clearly outlines 
early visual cortex which is expected to discriminate between stimuli of different sizes. 
}
\label{Fig:fig_w_clf}
\end{figure}

For both $\ell_1$ and hierarchical regularizations, one of the three vectors of
coefficients obtained for one fold and for the loss leading to sparser models are presented in Fig.~\ref{Fig:fig_w_clf}.
For $\ell_1$, the active voxels are scattered all over the brain, and for other
losses than the squared-loss the models selected tend not to be sparse. By contrast, the
tree $\ell_2$ regularization yields clearly delineated sparsity patterns located in the
visual areas of the brain. Like for the regression results, the highest
coefficients are obtained at scale 6 showing how spatially extended is the
brain region involved in the cognitive task. The symmetry of the pattern
at this scale is also particularly striking in the primary visual areas.
It also extends more anteriorly into the inferior temporal cortex,
known for high-level visual processing.

{%
\newcommand{\mc}[3]{\multicolumn{#1}{#2}{#3}}
\definecolor{tcA}{rgb}{0.627451,0.627451,0.643137}
\begin{table}
\begin{center}\footnotesize{
\begin{tabular}{|l|lll}\cline{1-4}
Loss function: & \mc{3}{c|}{Squared loss (``one-versus-all'')}\\\cline{1-4}
\mc{4}{>{\columncolor{tcA}}l}{×}\\\hline
\mc{1}{>{\columncolor{tcA}}l}{×}
& \mc{1}{c|}{Error (mean,std)}
& \mc{1}{c|}{P-value w.r.t. Tree $\ell_2$ ($\rho=1$)-ML}
& \mc{1}{c|}{Median fraction of non-zeros (\%)}\\\hline
Regularization: & \mc{1}{c|}{×} & \mc{1}{c|}{×} & \mc{1}{c|}{×}\\\hline
$\ell_2$ (Ridge) & \mc{1}{c|}{(29.2,\ 5.9)} & \mc{1}{c|}{0.004$^*$} & \mc{1}{c|}{100.00}\\\hline
$\ell_1$  & \mc{1}{c|}{(33.3,\ 6.8)} & \mc{1}{c|}{0.004$^*$} & \mc{1}{c|}{0.10}\\\hline
$\ell_1/\ell_2$ (Multi-task) & \mc{1}{c|}{(31.7,\ 9.5)} & \mc{1}{c|}{0.004$^*$} & \mc{1}{c|}{0.12}\\\hline
$\ell_1/\ell_\infty$ (Multi-task) & \mc{1}{c|}{(33.3,13.6)} & \mc{1}{c|}{0.009$^*$} & \mc{1}{c|}{0.22}\\\hline
Tree $\ell_2$ ($\rho=0.5$) & \mc{1}{c|}{(25.8,\ 9.2)} & \mc{1}{c|}{0.004$^*$} & \mc{1}{c|}{99.93}\\\hline
Tree $\ell_2$ ($\rho=1$) & \mc{1}{c|}{(25.0,\ 5.5)} & \mc{1}{c|}{0.027$^*$} & \mc{1}{c|}{10.08}\\\hline
Tree $\ell_2$ ($\rho=1.5$)  & \mc{1}{c|}{(24.2,\ 9.9)} & \mc{1}{c|}{\!\!\!0.130} & \mc{1}{c|}{0.05}\\\hline
Tree $\ell_\infty$ ($\rho=0.5$) & \mc{1}{c|}{(30.8,\ 8.8)} & \mc{1}{c|}{0.004$^*$} & \mc{1}{c|}{59.49}\\\hline
Tree $\ell_\infty$ ($\rho=1$) & \mc{1}{c|}{(24.2,\ 7.3)} & \mc{1}{c|}{\!\!\!0.058} & \mc{1}{c|}{1.21}\\\hline
Tree $\ell_\infty$ ($\rho=1.5$) & \mc{1}{c|}{(25.8,\ 10.7)} & \mc{1}{c|}{\!\!\!0.070} & \mc{1}{c|}{0.04}\\\hline
\mc{4}{>{\columncolor{tcA}}l}{×}\\\hline
Loss function: & \mc{3}{c|}{Logistic loss (``one-versus-all'')}\\\hline
\mc{4}{>{\columncolor{tcA}}l}{×}\\\hline
\mc{1}{>{\columncolor{tcA}}l}{×}
& \mc{1}{c|}{Error (mean,std)}
& \mc{1}{c|}{P-value w.r.t. Tree $\ell_2$ ($\rho=1$)-ML}
& \mc{1}{c|}{Median fraction of non-zeros (\%)}\\\hline
Regularization: & \mc{1}{c|}{×} & \mc{1}{c|}{×} & \mc{1}{c|}{×}\\\hline
$\ell_2$ (Ridge)  & \mc{1}{c|}{(25.0,\ 9.6)} & \mc{1}{c|}{0.008$^*$} & \mc{1}{c|}{100.00}\\\hline
$\ell_1$          & \mc{1}{c|}{(34.2,\ 15.9)} & \mc{1}{c|}{0.004$^*$} & \mc{1}{c|}{0.55}\\\hline
$\ell_1/\ell_2$ (Multi-task) & \mc{1}{c|}{(31.7,\ 8.6)} & \mc{1}{c|}{0.002$^*$} & \mc{1}{c|}{47.35}\\\hline
$\ell_1/\ell_\infty$ (Multi-task) & \mc{1}{c|}{(33.3,\ 10.4)} & \mc{1}{c|}{0.002$^*$} & \mc{1}{c|}{99.95}\\\hline
Tree $\ell_2$ ($\rho=0.5$) & \mc{1}{c|}{(25.0,\ 9.6)} & \mc{1}{c|}{0.007$^*$} & \mc{1}{c|}{99.93}\\\hline
Tree $\ell_2$ ($\rho=1$) & \mc{1}{c|}{(20.0,\ 11.2)} & \mc{1}{c|}{\!\!\!0.250} & \mc{1}{c|}{7.88}\\\hline
Tree $\ell_2$ ($\rho=1.5$) & \mc{1}{c|}{(18.3,\ 6.6)} & \mc{1}{c|}{\!\!\!0.500} & \mc{1}{c|}{0.06}\\\hline
Tree $\ell_\infty$ ($\rho=0.5$) & \mc{1}{c|}{(30.8,\ 10.4)} & \mc{1}{c|}{0.004$^*$} & \mc{1}{c|}{59.42}\\\hline
Tree $\ell_\infty$ ($\rho=1$) & \mc{1}{c|}{(24.2,\ 6.1)} & \mc{1}{c|}{0.035$^*$} & \mc{1}{c|}{0.60}\\\hline
Tree $\ell_\infty$ ($\rho=1.5$) & \mc{1}{c|}{(21.7,\ 8.9)} & \mc{1}{c|}{\!\!\!0.125} & \mc{1}{c|}{0.03}\\\hline
\mc{4}{>{\columncolor{tcA}}l}{×}\\\hline
Loss function: & \mc{3}{c|}{Multinomial logistic loss (ML)}\\\hline
\mc{4}{>{\columncolor{tcA}}l}{×}\\\hline
\mc{1}{>{\columncolor{tcA}}l}{×}
& \mc{1}{c|}{Error (mean,std)}
& \mc{1}{c|}{P-value w.r.t. Tree $\ell_2$ ($\rho=1$)-ML}
& \mc{1}{c|}{Median fraction of non-zeros (\%)}\\\hline
Regularization: & \mc{1}{c|}{×} & \mc{1}{c|}{×} & \mc{1}{c|}{×}\\\hline
$\ell_2$ (Ridge)  & \mc{1}{c|}{(24.2,\ 9.2)} & \mc{1}{c|}{0.035$^*$} & \mc{1}{c|}{100.00}\\\hline
$\ell_1$  & \mc{1}{c|}{(25.8,\ 12.0)} & \mc{1}{c|}{0.004$^*$} & \mc{1}{c|}{97.95}\\\hline
$\ell_1/\ell_2$ (Multi-task) & \mc{1}{c|}{(26.7,\ 7.6)} & \mc{1}{c|}{0.007$^*$} & \mc{1}{c|}{30.24}\\\hline
$\ell_1/\ell_\infty$ (Multi-task) & \mc{1}{c|}{(26.7,\ 11.6)} & \mc{1}{c|}{0.002$^*$} & \mc{1}{c|}{99.98}\\\hline
Tree $\ell_2$ ($\rho=0.5$) & \mc{1}{c|}{(22.5,\ 8.8)} & \mc{1}{c|}{\!\!\!0.070} & \mc{1}{c|}{83.06}\\\hline
Tree $\ell_2$ ($\rho=1$) & \mc{1}{c|}{(\textbf{16.7},\ \textbf{10.4})} & \mc{1}{c|}{-} & \mc{1}{c|}{4.87}\\\hline
Tree $\ell_2$ ($\rho=1.5$)  & \mc{1}{c|}{(18.3,\ 10.9)} & \mc{1}{c|}{\!\!\!0.445} & \mc{1}{c|}{0.02}\\\hline
Tree $\ell_\infty$ ($\rho=0.5$)  & \mc{1}{c|}{(26.7,\ 11.6)} & \mc{1}{c|}{0.015$^*$} & \mc{1}{c|}{48.82}\\\hline
Tree $\ell_\infty$ ($\rho=1$) & \mc{1}{c|}{(22.5,\ 13.0)} & \mc{1}{c|}{\!\!\!0.156} & \mc{1}{c|}{0.34}\\\hline
Tree $\ell_\infty$ ($\rho=1.5$) & \mc{1}{c|}{(21.7,\ 8.9)} & \mc{1}{c|}{\!\!\!0.460} & \mc{1}{c|}{0.05}\\\hline
\mc{4}{>{\columncolor{tcA}}l}{×}\\\hline
\mc{1}{>{\columncolor{tcA}}l}{×}
& \mc{1}{c|}{Error (mean,std)}
& \mc{1}{c|}{P-value w.r.t. Tree $\ell_2$ ($\rho=1$)-ML}
& \mc{1}{c|}{Median fraction of non-zeros (\%)}\\\hline
Greedy & \mc{1}{c|}{(21.6,\ 14.5)} & \mc{1}{c|}{0.001$^*$} & \mc{1}{c|}{0.01}\\\hline
\end{tabular}}
\end{center}
\caption{Prediction results obtained on fMRI data (see text) for the multi-class classification setting.
From the left, the first column contains the mean and standard deviation of the test error (percentage of misclassification),
computed over leave-one-subject-out folds.
The best performance is obtained with the hierarchical $\ell_2$ penalization 
($\rho=1$) constructed from the Ward tree,
coupled with the multinomial logistic loss function.
Methods with performance significantly worse than this combination is assessed by Wilcoxon two-sample paired
signed rank tests (The superscript $^*$ indicates a rejection at $5\%$). Levels of sparsity reported are in the augmented space whenever it is used.}
\label{Tab:res_sizes_inter_cl}
\end{table}
}

\section{Conclusion}

In this article, we introduced a hierarchically structured regularization, which
takes into account the spatial and multi-scale structure of fMRI data. 
This approach copes with inter-subject variability in a similar way
as feature agglomeration, by averaging neighboring voxels.
Although alternative agglomeration strategies do exist, we simply used the criterion which appears as the most natural, Ward's clustering,
and which builds parcels with little variance.

Results on a real dataset show that the proposed algorithm is a promising tool for
mining fMRI data. It yields similar or higher prediction accuracy than reference methods,
and the map of weights it obtains exhibit a cluster-like structure. It makes
them easily readable compared to the overly sparse patterns found by classical
sparsity-promoting approaches.

For the regression problem, both the greedy method from~\cite{michel2010} and the
proposed algorithm yield better results than unstructured and non-hierarchical
regularizations, whereas in the classification setting, our approach leads to the best performance.
Moreover, our proposed methods enjoy the different benefits from convex optimization.
In particular, while the greedy algorithm relies on a two-step approach that may
be far from optimal, the hierarchical regularization induces simultaneously the
selection of the optimal parcellation and the construction of the optimal predictive
model, given the initial hierarchical clustering of the voxels. Moreover, convex methods
yield predictors that are essentially stable with respect to perturbations of the design or the initial clustering, which is typically not the case of greedy methods.
It is important to distinguish here the stability of the predictors from that of the only learned map $\wb$, which could be enforced via a squared $\ell_2$-norm regularization.

Finally, it should be mentioned that the performance achieved by this approach in
inter-subject problems suggests that it could potentially be used successfully  for medical diagnosis,
in a context where brain images --not necessarily functional images-- are used to classify individuals
into diseased or control population. Indeed, for difficult problems of that sort, where the reliability
of the diagnostic is essential, the stability of models obtained from convex formulations
and the interpretability of sparse and localized solutions are quite relevant to provide a credible diagnostic.

\section*{Acknowledgments}
The authors acknowledge support from the ANR grants
ViMAGINE ANR-08-BLAN-0250-02 and ANR 2010-Blan-0126-01 ``IRMGroup''.
The project was also partially supported by a grant from the European Research Council (SIERRA Project).

\bibliographystyle{siam}
\bibliography{sc,main_bibliography}

\end{document}